\documentclass[letterpaper]{article} 
\usepackage{aaai25}  
\usepackage{times}  
\usepackage{helvet}  
\usepackage{courier}  
\usepackage[hyphens]{url}  
\usepackage{graphicx} 
\usepackage{natbib}  
\usepackage{caption} 
\usepackage{algorithm}
\usepackage{algorithmicx}
\usepackage{algpseudocode} 
\usepackage{multirow} 
\usepackage{subfigure}
\usepackage{amsfonts}

\usepackage{algpseudocode}
\usepackage{amsmath}
\usepackage{tcolorbox}

\usepackage{multirow}
\usepackage{diagbox}
\usepackage{booktabs}
\usepackage{colortbl}
\usepackage{amssymb}
\usepackage{colortbl}
\usepackage{arydshln}

\usepackage{newfloat}
\usepackage{listings}
\DeclareCaptionStyle{ruled}{labelfont=normalfont,labelsep=colon,strut=off} 
\floatstyle{ruled}
\newfloat{listing}{tb}{lst}{}
\floatname{listing}{Listing}
\title{\textit{Utilize the Flow before Stepping into the Same River Twice}: Certainty Represented Knowledge Flow for Refusal-Aware Instruction Tuning}
\author {
    Runchuan Zhu\textsuperscript{$\clubsuit$}\textsuperscript{$\spadesuit$}\equalcontrib,
    Zhipeng Ma\textsuperscript{$\heartsuit$}\equalcontrib,
    Jiang Wu\textsuperscript{$\spadesuit$}\equalcontrib\thanks{Project lead.}, \\
    Junyuan Gao\textsuperscript{$\diamondsuit$}\textsuperscript{$\spadesuit$},
    Jiaqi Wang\textsuperscript{$\spadesuit$},
    Dahua Lin\textsuperscript{$\spadesuit$},
    Conghui He\textsuperscript{$\spadesuit$}\thanks{Corresponding author (heconghui@pjlab.org.cn).}
}
\affiliations {
    \textsuperscript{$\spadesuit$}Shanghai Artificial Intelligence Laboratory\\
    \textsuperscript{$\clubsuit$}Peking University\\
    \textsuperscript{$\heartsuit$}Southwest Jiaotong University\\
    \textsuperscript{$\diamondsuit$}Hangzhou Institute for Advanced Study, University of Chinese Academy of Sciences\\
    \{zhurunchuan, wujiang, gaojuanyuan, wangjiaqi, lindahua, heconghui\}@pjlab.org.cn, \\
    mazhipeng1024@my.swjtu.edu.cn
}

\usepackage{bibentry}

\begin{document}

\maketitle

\begin{abstract}
Refusal-Aware Instruction Tuning (RAIT) enables Large Language Models (LLMs) to refuse to answer unknown questions. By modifying responses of unknown questions in the training data to refusal responses such as ``I don't know", RAIT enhances the reliability of LLMs and reduces their hallucination.
Generally, RAIT modifies training samples based on the correctness of the initial LLM's response. However, this crude approach can cause LLMs to excessively refuse answering questions they could have correctly answered, the problem we call over-refusal.
In this paper, we explore two primary causes of over-refusal:
\textbf{\textit{Static conflict}} occurs when similar samples within the LLM’s feature space receive differing supervision signals (original vs. modified ``I don't know"). 
\textbf{\textit{Dynamic conflict}} arises as the LLM's evolving knowledge during SFT enables it to answer previously unanswerable questions, but the now-answerable training samples still retain the original ``I don't know" supervision signals from the initial LLM state, leading to inconsistencies.
These conflicts cause the trained LLM to misclassify known questions as unknown, resulting in over-refusal.
To address this issue, we introduce Certainty Represented Knowledge Flow for Refusal-Aware Instructions Tuning (CRaFT). CRaFT centers on two main contributions: First, we additionally incorporate response certainty to selectively filter and modify data, reducing static conflicts. Second, we implement preliminary rehearsal training to characterize changes in the LLM's knowledge state, which helps mitigate dynamic conflicts during the fine-tuning process. 
We conducted extensive experiments on open-ended question answering and multiple-choice question task. Experiment results show that CRaFT can improve LLM's overall performance during the RAIT process.
\end{abstract}

\begin{links}
    \link{Code \& Data}{https://github.com/opendatalab/CRaFT}
\end{links}

\section{Introduction}

Recently, Large Language Model (LLM) technology has made significant progress, becoming an important milestone towards AGI \cite{achiam2023gpt, dubey2024llama, touvron2023llama, yang2024qwen2}. However, current LLMs often output fabricated information, which is referred to as hallucinations \cite{ji2023survey}. This phenomenon severely limits the usefulness and reliability of LLMs. An important reason for the occurrence of hallucinations is that when exposed to questions beyond their internal knowledge (i.e., unknown questions), LLMs may forcefully generate responses \cite{kang2024unfamiliar}. Ideally, the reliable LLM should actively refuse to answer questions it doesn't know to avoid incorrect responses \cite{know_your_limits,survey_honesty}. Recent studies have shown \cite{alignment_for_honesty, R_Tuning, rejection_improves, cheng2024can, rejection_improves, Bai_Zhang_Tao_Wu_Wang_Xu_2023, cheng2024can} that Refusal-Aware Instruction Tuning (RAIT) can enable LLMs to refuse answering questions beyond their knowledge.

\begin{figure}[t]
    \centering
    \vspace{-5pt}
    \includegraphics[width=3.2in]{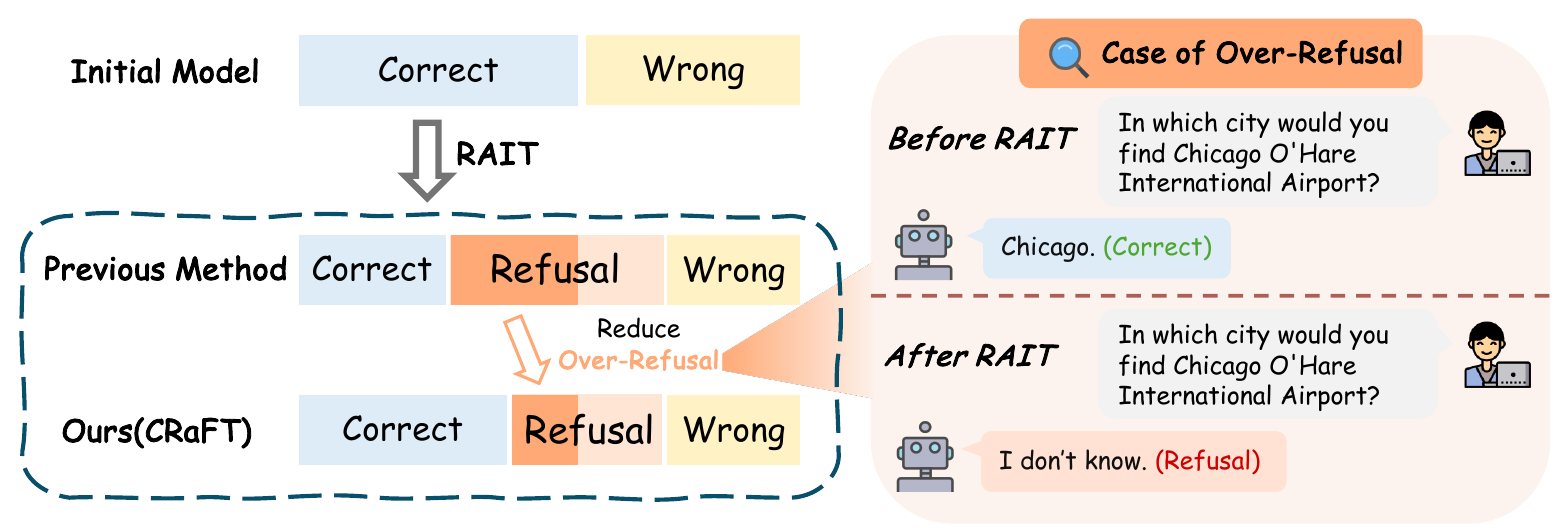}
    \caption{Previous RAIT methods resulted in significant over-refusal, while our CRaFT mitigates this issue, enhancing the LLM's reliability and helpfulness.}
    \label{fig:over_refusal}
    \vspace{-5pt}
\end{figure}

Generally, the RAIT process cab be described as follows: the initial LLM is prompted to answer all questions in the train set $D_{\text{src}}$. Based on the correctness of the responses, the samples are divided into two groups. Correct responses are considered known knowledge and are labeled as vanilla samples \( D_{\text{van}} \), whose answers remain unchanged. Incorrect responses are considered unknown knowledge and are replaced with "I don't know", forming the IdK samples \( D_{\text{idk}} \). The combined RAIT data, \( D_{\text{rait}} = D_{\text{van}} \cup D_{\text{idk}} \), is used to fine-tune the LLM, enhancing its ability to refuse unknown questions. For simplicity, we refer to this original form of the RAIT method as Cor-RAIT.

\begin{figure*}[t]
    \centering
    \vspace{-5pt}
    \includegraphics[width=6.0in]{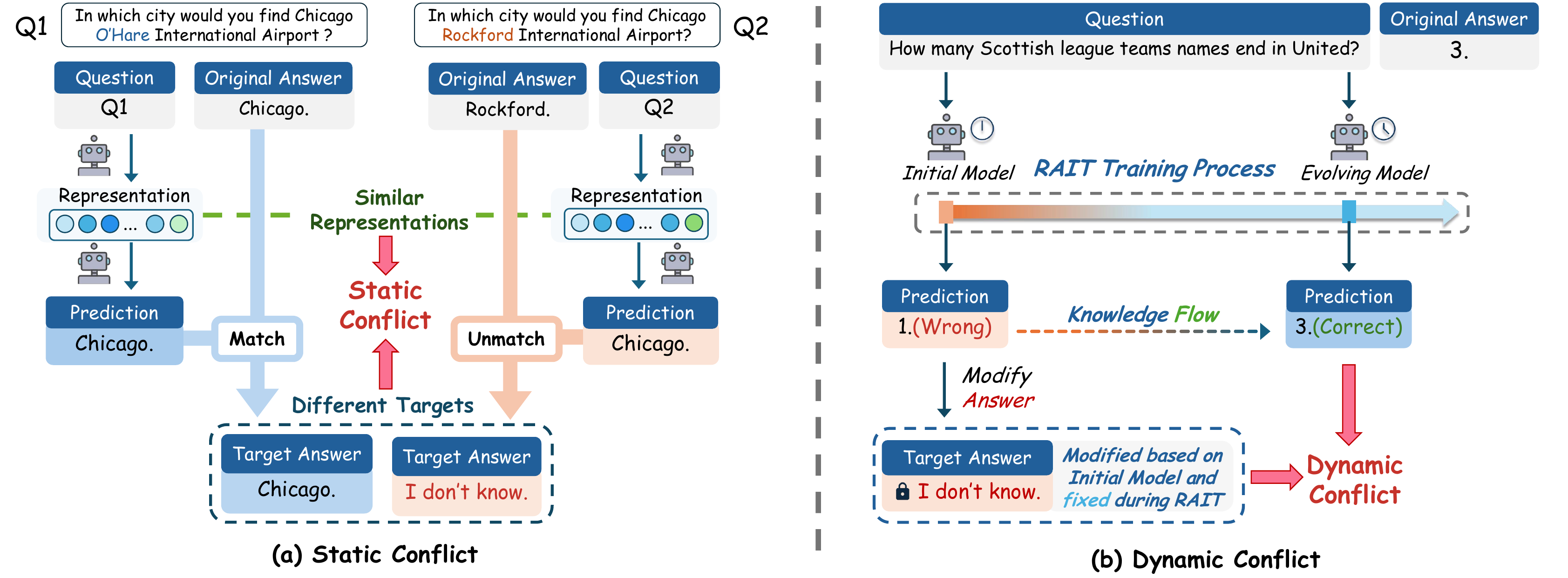}
     \caption{Two causes of over-refusal: (a) \textit{Static conflict} means the similar samples in the LLM's feature space being assigned different labels (original vs. modified ``I don't know").    
     (b) \textit{Dynamic conflict} arises since the LLM's knowledge state evolves during SFT, turning initally unknown questions to knowns, while the target answer remains IdK.
     These conflicts cause the trained LLM to misclassify known questions as unknown, resulting in over-refusal.}
    \label{fig:static_and_dynamic_conflicts}
\end{figure*}

However, \cite{cheng2024can} shows that Cor-RAIT causes the fine-tuned LLM to refuse some questions that could have been answered correctly. Experiments reveal a significant accuracy drop after Cor-RAIT: on the TriviaQA dataset \cite{triviaqa}, accuracy falls from 45.05\% to 28.57\%, and on the Natural Questions dataset \cite{nq}, it drops from 24.65\% to 15.93\%. We refer to this phenomenon as \textbf{over-refusal}, as shown in Figure \ref{fig:over_refusal}.

In addressing the over-refusal brought by Cor-RAIT, we identified two primary causes as shown in Figure 2.
\textbf{(1) Static Conflict}: In the LLM representation space, two closely located samples might be assigned to \(D_{\text{van}}\) and \(D_{\text{idk}}\) under the Cor-RAIT framework. As illustrated by t-SNE in Figure 3(a), significant intersections exist between \(D_{\text{van}}\) and \(D_{\text{idk}}\), complicating their differentiation. These similar samples provide conflict supervision during training, impairing the LLM's ability to distinguish between known and unknown questions, resulting in over-refusal.
\textbf{(2) Dynamic Conflict}: This arises from overlooking the dynamic shifts in LLM's knowledge state during training. Research \cite{ren2024learning_or_align, ren2024learning_dynamics_of_FT, xu2024parenting} shows that the knowledge state of LLMs changes during Supervised Fine-Tuning (SFT), with questions potentially shifting from unknown to known and vice versa.  This phenomenon is reminiscent of Heraclitus' saying, ``\textit{no man ever steps in the same river twice}." However, current methods use static RAIT data reflecting the initial LLM's knowledge state throughout SFT, which ignores these changes. This oversight leads to conflicts between the RAIT data and the LLM's evolving knowledge, resulting inefficient training and over-refusal.

To address the two problems above, we propose Certainty Represented Knowledge Flow for Refusal-Aware Instructions Construction (CRaFT). Our approach consists of two stages. 
\textbf{Stage 1: Querying the Knowledge State and Flow of the LLM.} First, we probe the initial LLM's knowledge state. Unlike Cor-RAIT, we incorporate response certainty alongside correctness, effectively alleviating the \textit{static} conflict between the supervision signals in \( D_{\text{van}} \) and \( D_{\text{idk}} \). To capture the LLM's dynamic knowledge changes during training, we introduce a rehearsal training mechanism. This fine-tunes the LLM with data samples that align closely with its internal knowledge, without introducing new knowledge \cite{ren2024learning_or_align,kang2024unfamiliar}. This approach allows us to observe the LLM’s natural knowledge adjustments. The differences between the fine-tuned and initial LLMs reveal the knowledge flow during training, helping to identify and resolve \textit{dynamic} conflicts.
\textbf{Stage 2: Refusal-aware instructions construction and tuning}. By considering both the static knowledge state and dynamic knowledge flow, we filter out vanilla and IdK samples from RAIT data, reducing conflicts. We then fine-tune the initial LLM with the refined data, improving overall performance.

In conducting our experimental analysis, we sought a well-founded metric within current research. Existing methods have notable limitations, either proposing multiple metrics that are hard to optimize simultaneously or relying on inherently flawed metrics, as demonstrated by our counterexamples. Consequently, we examined these shortcomings and introduced a singular and comprehensive metric: \textbf{Truthful Helpfulness Score(THS)}.

Overall, our main contributions are as follows:

\begin{itemize}

 \item We conducted the in-depth analysis of static and dynamic conflicts in existing correctness-based RAIT data, revealing that they cause the trained LLMs' mis-classification of known and unknown questions, leading to the issue of over-refusal in current RAIT methods.

 \item To address static and dynamic conflicts, we introduce CRaFT: it reduces static conflicts by incorporating certainty alongside correctness during RAIT data construction, and mitigates dynamic conflicts through rehearsal training to capture knowledge flow trends. Extensive experiments demonstrate that CRaFT alleviates over-refusal and improves overall LLM performance.

 \item We analyze the shortcomings of existing refusal-aware metrics and introduce the Truthful Helpfulness Score (THS), which balances reliability and helpfulness for a comprehensive evaluation of LLM performance.

 \begin{figure}[t]
    \centering
    \vspace{-5pt}
    \includegraphics[width=3in]{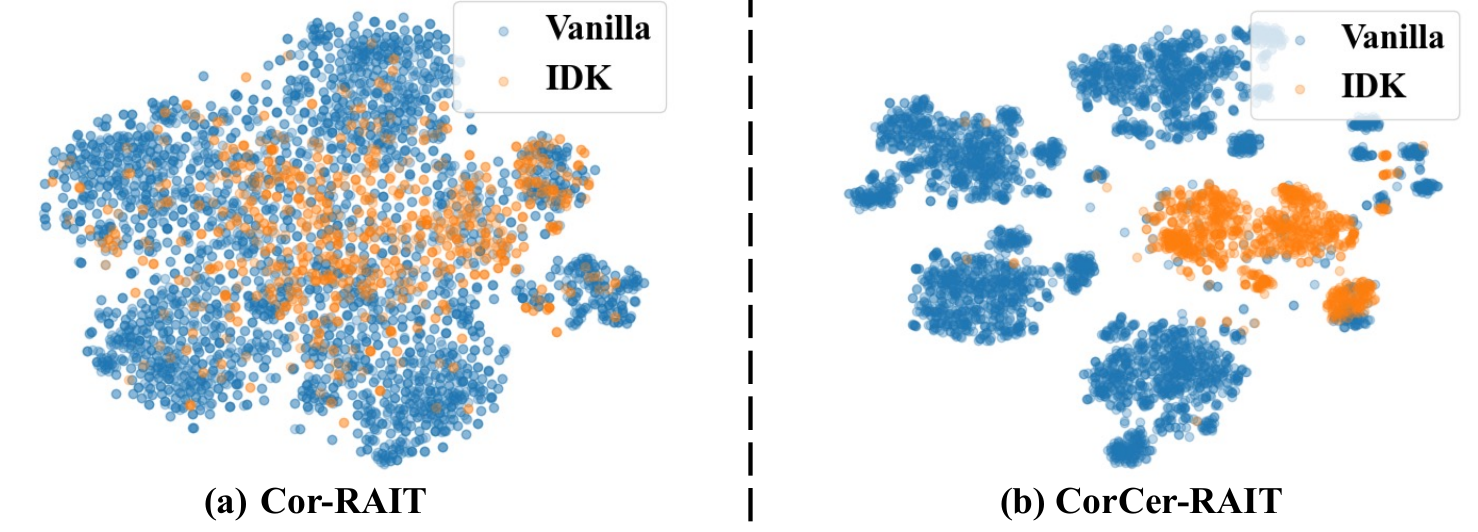}
    \caption{t-SNE visualization of the LLM feature space}
    \label{fig:t_sne}
\end{figure}

 \end{itemize}
\section{Related Work}

\subsection{Mitigating Hallucinations of LLMs}
Researchers have developed various methods to mitigate LLM hallucinations, including data augmentation \cite{neeman2022disentqa}, improved decoding strategies \cite{holtzman2019curious, chuang2023dola}, external knowledge integration \cite{karpukhin2020dense}, knowledge editing \cite{zhang2024truthx, li2024inference}, and honesty alignment \cite{R_Tuning, rejection_improves, bai2024efficient}. Unlike traditional correction methods, honesty alignment encourages models to say ``I don't know" for unknown questions.

\subsection{Refusal-Aware Instruction Tuning}
RAIT is the supervised technique that improves LLMs' responses by training LLMs to directly respond with ``I don't know" to unknown questions. R-Tuning \cite{R_Tuning} identifies these questions by having the LLM answer each once and verifying response accuracy. In \cite{alignment_for_honesty}, the LLM answers the same question multiple times, with the target answer adjusted based on the correctness ratio. \cite{wan2024mitigating} uses a knowledge-based verification mechanism to ensure consistency with trusted external sources, enhancing refusal accuracy and preventing misinformation.

\section{Over-Refusal: Analysis and Insights}

\subsection{Refusal-Aware Instruction Tuning}

Given the initial LLM \( \mathcal{M}_0 \) and the instruction dataset \( D_{\text{src}} \) of question-answer pairs \( x = (q, a) \), we modify \( D_{\text{src}} \) to construct \( D_{\text{rait}} \), consisting of pairs \( (q, a_{\text{rait}}) \). \( D_{\text{rait}} \) is then used for SFT on \( \mathcal{M}_0 \), resulting in a new LLM capable of declining unknown questions, a process called Refusal-Aware Instruction Tuning (RAIT). Existing studies \cite{R_Tuning, alignment_for_honesty, cheng2024can} use \( \mathcal{M}_0 \) to infer and assess the correctness of questions in \( D_{\text{src}} \), denoted as \( \mu \). As shown in Figure 4(a), the correctness threshold \( \tau_{\mu} \) is first defined. If \( \mu < \tau_{\mu} \), the answer is changed to "I don't know" and assigned to the IdK subset \( D_{\text{idk}} \). If \( \mu \ge \tau_{\mu} \), the original answer remains, and the pair is assigned to the vanilla subset \( D_{\text{van}} \). The resulting RAIT data is \( D_{\text{rait}} = D_{\text{van}} \cup D_{\text{idk}} \), and this correctness-based RAIT is called Cor-RAIT. 

However, LLMs exhibited significant over-refusal after Cor-RAIT, as shown in Figure \ref{fig:over_refusal}. Subsequent sections of this chapter will analyze the causes and offer practical insights.

\subsection{Static Conflicts in Cor-RAIT}

During the Cor-RAIT process, LLMs learn to reject unknown samples by supervisions from \(D_{\text{idk}}\). Our insight is that if the Cor-RAIT dataset contains vanilla and IdK samples that are closely positioned in the LLM's representation space, the trained LLM may mistakenly classify similar vanilla samples as IdK samples, causing over-refusals. To verify this, we analyzed the sample distributions of \(D_{\text{van}}\) and \(D_{\text{idk}}\). We extract latent representation of each question from the last hidden layer of the LLM. Then, t-SNE is adopted to visualize sample representations.
Figure 3(a) displays the distributions of samples from the test split of MMLU dataset \cite{MMLU} in the LLaMA-3-8B-instruct \cite{dubey2024llama} representation space, where IdK and vanilla samples have significant intersections. \footnote{More experiments involving additional datasets and LLMs are provided in Appendix A.1.}

Furthermore, we introduce the \textbf{C}onflict \textbf{R}ate for \textbf{S}imilar \textbf{S}amples (CRSS) to quantitatively assess conflicts in supervision signals among similar samples in the RAIT dataset. For each sample \(x_i\) in \(D_{\text{idk}}\), we compute the cosine similarity between its question representation \(r_i\) and the question representation \(r_j\) of each sample \(x_j\) in \(D_{\text{van}}\).  We identify and record the highest similarity value obtained. If this value exceeds the predefined similarity threshold \(\tau_{\text{sim}}\), we record the occurrence. The CRSS is then calculated as:
\[
\text{CRSS} = \frac{\sum_{x_i \in D_{\text{idk}}} \mathbf{1}\left(\max_{x_j \in D_{\text{van}}} \text{cos}(r_i, r_j) > \tau_{\text{sim}}\right)}{|D_{\text{idk}}|}
\]
Therefore, the higher CRSS indicates more conflicting similar sample pairs, potentially leading to over-refusal.
We computed the CRSS for Cor-RAIT, as shown in Figure 5. The results show that at \(\tau_{\text{sim}} = 0.97\), CRSS reaches significant levels across various LLM and dataset combinations, supporting earlier t-SNE findings\footnote{We show more results and details in Appendix A.4.}.

The above analysis reveals that Cor-RAIT generates numerous similar sample pairs between \(D_{\text{van}}\) and \(D_{\text{idk}}\), resulting in conflicting supervision signals which leads to over-refusal. We term this \textbf{static conflict} to distinguish it from another conflict type discussed later.

\begin{figure}[t]
    \centering
    \vspace{-5pt}
    \includegraphics[width=3in]{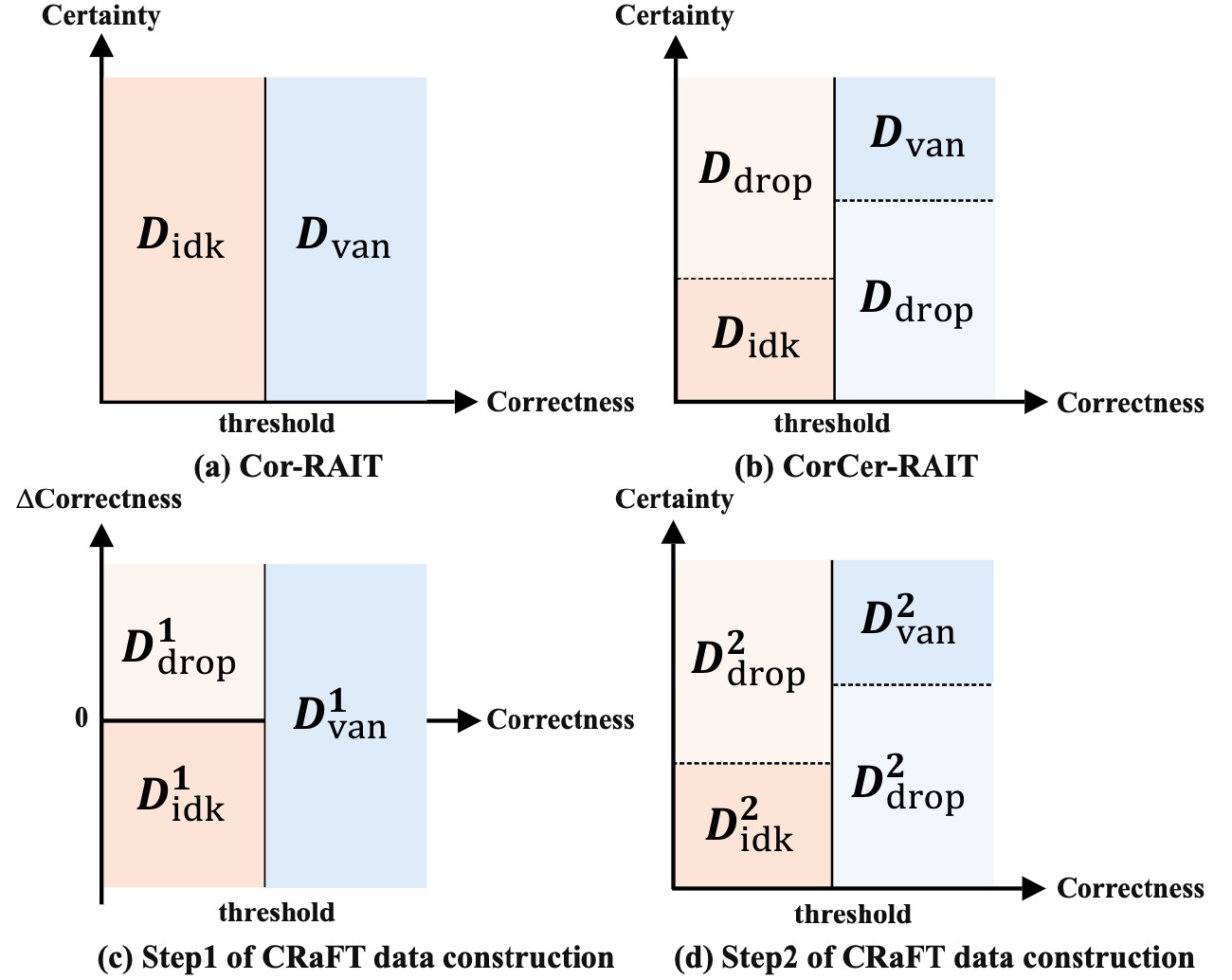}
    \caption{RAIT Data Construction of Cor-RAIT, CorCer-RAIT and CRaFT. Cor-RAIT partitions data based on accuracy \(\mu\) and a threshold \(\tau_{\mu}\). For CorCer-RAIT, \(D_{\text{van}}\) is derived from samples with accuracy exceeding the threshold and the highest certainty, while \(D_{\text{idk}}\) consists of samples with accuracy below the threshold and the lowest certainty. CRaFT employs a two-stage process: in the first stage, data with \(\Delta\mu > 0\) is excluded through knowledge queries; the second stage follows the same procedure as CorCer-RAIT.}
    \vspace{-5pt}
    \label{figure:our_method}
\end{figure}

\subsection{Certainty Mitigates the Static Conflicts}

We conducted a theoretical analysis \footnote{detailed proof in Appendix A.2.} establishing a weak (non-differentiable) link between the LLM's feature and the response correctness \(\mu\) for the specific question \(q\). This weak correlation causes highly similar samples being categorized into \(D_{\text{van}}\) and \(D_{\text{idk}}\) respectively.
To mitigate this, we propose incorporating a robust indicator variable aligned with \textit{correctness} to select and construct the RAIT data. This variable should  ensure that similar samples share comparable values, reducing the above misclassification.
We suggest adopting the \textit{certainty} \cite{jiang2023uncertainty} of the LLM's response as the indicator variable. Our theoretical analysis shows that certainty meets the above requirements.\footnote{Section 4.2 discusses various methods for representing LLM response certainty. Appendix A.3 uses entropy as a measure, but our findings extend to other methods as well.}

To incorporate correctness and certainty into the RAIT data selection, we developed the CorCer-RAIT framework, as shown in Figure 4(b). We visualized the sample distribution in \(D_{\text{van}}\) and \(D_{\text{idk}}\) using t-SNE in the LLM representation space, as shown in Figure 3(b), which shows a significant decrease in the overlap between \(D_{\text{van}}\) and \(D_{\text{idk}}\) compared to the Cor-RAIT in Figure 3(a). Furthermore, we calculated the CRSS for both methods, as shown in Figure 5, highlighting substantial reductions in CorCer-RAIT over Cor-RAIT. Therefore, the joint use of correctness and certainty effectively alleviates the \textit{static} conflict between the supervision signals in $D_{\text{van}}$ and $D_{\text{idk}}$.

\begin{figure}[th]
    \centering
    \vspace{-5pt}
    \includegraphics[width=3.3in]{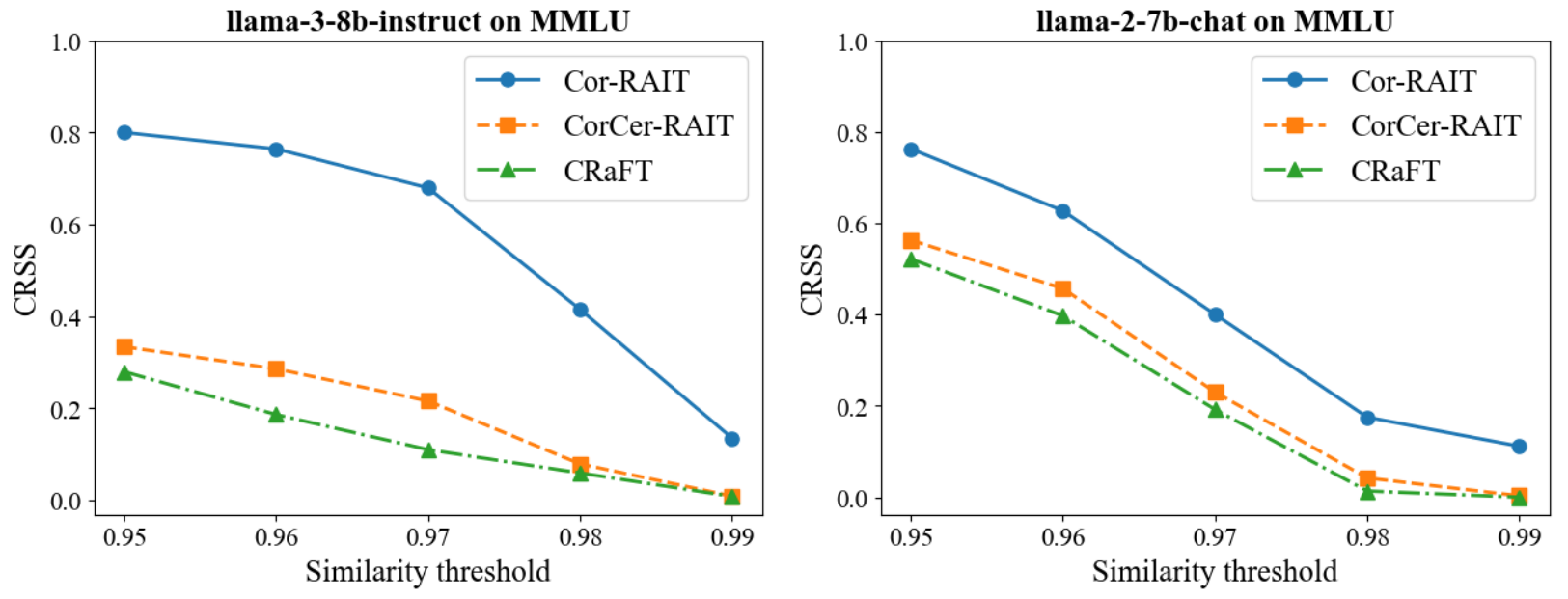}
    \caption{CRSS of Different RAIT Samples.}
    \vspace{-5pt}
\end{figure}

\subsection{Knowledge Flow and Dynamic Conflict}

Research \cite{ren2024learning_or_align,gekhman2024does,ren2024learning_dynamics_of_FT} reveals that the knowledge state of LLMs evolves during the SFT process. The phenomenon, which we refer to as ``knowledge flow", can cause previously incorrectly answered questions to become correct ones during SFT.
Despite this dynamic evolutions, the target answer of the training data remains static during the RAIt, which reflects the knowledge state of the initial LLM but ignores subsequent changes. We term it as \textbf{dynamic conflict}, which significantly contributes to the over-refusal in Cor-RAIT.

We select the data with the highest correctness and certainty for SFT, a process we refer to as \emph{rehearsal training}\footnote{Details are provided in Appendix B.2.}. \textit{Rehearsal training} is designed to capture the LLM's natural knowledge flow during SFT. Experiments on the MMLU dataset \cite{MMLU} and LLaMA3-8B-Instruct demonstrated that the correctness of \textbf{69\%} of samples, initially below 0.5, improved, thereby validating the aforementioned analysis on dynamic conflict. Additional experimental results are provided in Appendix A.5.

\section{Methodology}

\subsection{Overview}

Based on Section 3, we propose the Certainty Represented Knowledge Flow for Refusal-Aware Instructions Construction (CRaFT) to solve the over-refusal problem. CRaFT contains two stages, as shown in Figure 6.

\begin{figure}[t]
    \centering
    \vspace{-5pt}
    \includegraphics[width=3.2in]{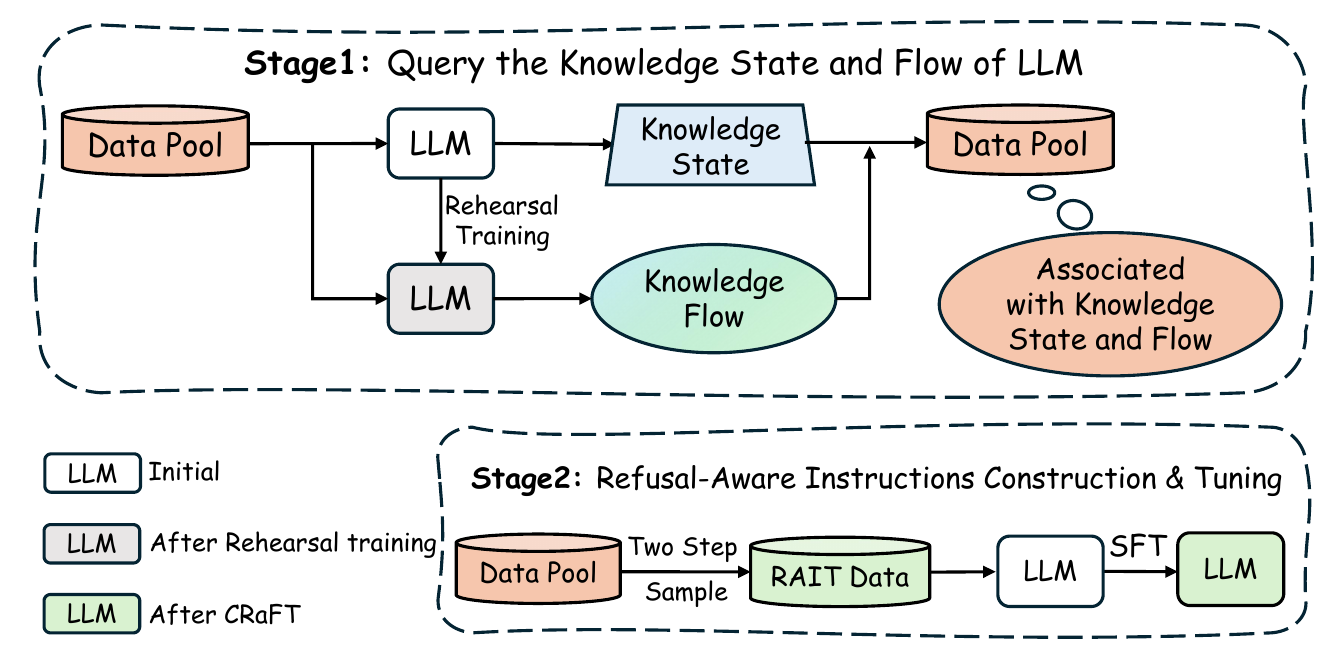}
    \caption{The Framework of CRaFT: Stage 1 queries knowledge state and flow, while Stage 2 constructs RAI data and tunes.}
    \label{figure:framework}
    \vspace{-5pt}
\end{figure}

\subsubsection{Stage 1: Query the Knowledge State and Flow of LLM}

The output of stage one is the knowledge state and flow indicators of the model. First, we perform a knowledge state query to obtain the correctness and certainty of the model's responses to the samples in the source dataset. Next, we conduct rehearsal training on the model, resulting in the perturbed version. By comparing the knowledge states before and after perturbation, we derive the indicators of knowledge flow during the supervised fine-tuning process.

\subsubsection{Stage 2: Refusal-Aware Instructions Construction and Tuning}

Using the knowledge state and flow from Stage 1, we select suitable samples from \(D_{\text{src}}\) to construct the RAIT data, which is is used to fine-tune the initial model.

\subsection{Query the Knowledge State and Flow of LLM}

\subsubsection{Knowledge State Query}
The input for knowledge state query consists of the LLM ( $\mathcal{M}_{0}$ or \(\mathcal{\widetilde{M}}\)) and  \(D_{\text{src}}\). The output is the LLM's correctness and certainty for each sample in \(D_{\text{src}}\), represented as \(\{\mu_0=Cor(\mathcal{M}, x_i),\sigma_0=Cer(\mathcal{M}, x_i) | x_i \in D_{\text{src}} \}\), which indicate the LLM's knowledge state. Our research focuses on the Multiple-Choice Question Answering (MCQA) and Open-ended Questions Answer (OEQA) tasks, which correspond to different methods of knowledge state query.

In the MCQA task, for a given model \(\mathcal{M}\) and question \(q\), the possible answers are included in \(O = \{A, B, C, D\}\). We obtain the token probability of \(\hat{a}\), denoted as \(p(\hat{a}|q, \mathcal{M})\), where \(\hat{a} \in O\).
We use the probability of the target answer token to represent correctness. Certainty is calculated through negative entropy. The corresponding formulas are:
$$
\begin{array}{l} 
  Cor(\mathcal{M}, x_i) = p(x_i.a|x_i.q, \mathcal{M}),  \\  
  Cer(\mathcal{M}, x_i) = -\sum_{\hat{a} \in O} p(\hat{a}|x_i.q, \mathcal{M}) \log(p(\hat{a}|x_i.q, \mathcal{M}))
\end{array} 
$$

In the OEQA task, following \cite{alignment_for_honesty,cheng2024can}, given a sample \(x_i\), the LLM \(\mathcal{M}\) performs inference on \(x_i.q\) and generates responses \(N\) times (with \(N=10\)). The generated responses \(\{\hat{a}_0, \dots, \hat{a}_{N-1}\}\) are denoted as \(\hat{A}_i\). The generation process is carried out with a temperature of 1.0 and sampling enabled (\texttt{do\_sample=True}).
$$
\begin{array}{l} 
Cor(\mathcal{M}, x_i) = \frac{1}{N} \sum_{\hat{a}_j \in \hat{A_i}} \mathbf{1}(\hat{a}_j = x_i.a) \\ 
Cer(\mathcal{M}, x_i) = \frac{1}{N(N-1)} \sum_{\substack{\hat{a}_j, \hat{a}_k \in \hat{A_i}, j \neq k}} \text{cos}(E(\hat{a}_j), E(\hat{a}_k))
\end{array} 
$$

Correctness is obtained through exact match across the \(N\) responses, calculating the proportion of accurate answers. Certainty is evaluated using a pretrained SentenceTransformer model \footnote{\url{https://huggingface.co/sentence-transformers/all-MiniLM-L6-v2}} to encode each response \(\hat{a}_j\) into embedding \(E(\hat{a}_j)\), and the average similarity is computed between these embeddings (excluding diagonal elements).
 The correctness values range from $[0, 1]$. In MCQA task, certainty ranges from [$-log|O|$, 0], and for OEQA, from $[0, 1]$. More details about knowledge state query are in Appendix B.1.

\subsubsection{Rehearsal Training and Knowledge Flow}

During rehearsal training, we select high-certainty and high-correctness samples from \(D_{\text{src}}\) to fine-tuning \(\mathcal{M}_{0}\). \(\mathcal{\widetilde{M}}\) is obtained after fine-tuning. In the same way, we assess the perturbed LLM's knowledge state by performing another knowledge state query, yielding correctness and certainty for each QA pair in \(D_{\text{src}}\): \(\{\tilde{\mu} = Cor(\mathcal{\widetilde{M}}, x_i), \tilde{\sigma} = Cer(\mathcal{\widetilde{M}}, x_i)| x_i \in D_{\text{src}} \}\). The knowledge flow from the original \(\mathcal{M}_{0}\) to the perturbed \(\mathcal{\widetilde{M}}\) is quantified as:
\[
\Delta\mu = Cor(\mathcal{\widetilde{M}}) - Cer(\mathcal{M}_{0})
\]
\[
\Delta\sigma = Cor(\mathcal{\widetilde{M}}) - Cer(\mathcal{M}_{0})
\]

Rehearsal training sample selection prioritizes those with the highest correctness and certainty. This insight is supported by \cite{ren2024learning_or_align,kang2024unfamiliar,gekhman2024does}, which indicates that LLMs primarily refine and activate existing knowledge rather than acquire new knowledge during SFT. We align the rehearsal training with the LLM's internal knowledge state, ensuring a more natural and effective knowledge flow during the SFT process.
More details about rehearsal training are in Appendix \ref{B2}.

\subsection{Refusal-Aware Instructions Constuction and Tuning}
Unlike Cor-RAIT, which selects RAIT samples solely based on correctness, our approach leverages four parameters \(\mu\), \(\sigma\), \(\Delta\mu\), and \(\Delta\sigma\) to characterize both the knowledge state and flow of $\mathcal{M}_{0}$. The challenge lies in making informed sample selections across these four dimensions. We propose a two-step heuristic method outlined in Algorithm 1.

\subsubsection{Step 1}
As shown in Figure 4(c), we first filter the training sample \(D_{\text{src}}\) on the \(\mu\) and \(\Delta\mu\) plane. Setting a correctness threshold \(\tau_{\mu}\), we define the vanilla candidate set \(D_{\text{van}}^1 = \{x_i|x_i.\mu\ge\tau_\mu\}\). For IdK candidates, unlike Cor-RAIT, we select \(D_{\text{idk}}^1 = \{x_j|x_j.\mu<\tau_\mu~and~x_j.\Delta\mu < 0\}\). Samples in \(D_{\text{drop}}^1 = \{x_k | x_k.\mu < \tau_\mu~and~x_k.\Delta\mu\ge0\}\) are discarded because their correctness is actively increasing during SFT, shifting from unknown to known, which could lead to dynamic conflicts.

\subsubsection{Step 2}
As shown in Figure 4(d), we sort both \(D_{\text{van}}^1\) and \(D_{\text{idk}}^1\) by certainty $\sigma$. From \(D_{\text{van}}^1\), we select the top \(N_{\text{van}}\) samples as final vanilla samples \(D_{\text{van}}^2\), and the bottom \(N_{\text{idk}}\) samples as IdK candidates of \(D_{\text{idk}}^2\), whose answers are then modified to ``I don't know". The samples in \(D_{\text{drop}}^2\) are discarded. The final RAIT data \(D_{\text{rait}} = D_{\text{van}}^2 \cup D_{\text{idk}}^2\).

\begin{algorithm}[t]
\caption{RAIT Data Construction Process}
\label{alg:RAI_data}
\textbf{Input:}{~$D_{\text{src}} = \{x_{0}, x_{1}, ..., x_{N}\}$}, $\tau_\mu$, $N_{\text{van}}$, $N_{\text{idk}}$ \\
\textbf{Output:}{~$D_{rait}\subseteq D_{\text{src}}$}
\begin{algorithmic}[1]
\fontsize{9}{8}
\State $D_{\text{van}}^1 = \{x_i \, | \, x_i \in D_{\text{src}},  x_i.\mu \geq \tau_\mu\}$
\State $D_{\text{idk}}^1 = \{x_j \, | \, x_j \in D_{\text{src}}, x_j.\mu < \tau_\mu \text{ and } x_j.\Delta\mu < 0\}$
\State $D_{\text{\text{van}}}^1 = \text{sort}(D_{\text{van}}^1, \text{key}=\sigma, \text{order=descend})$
\State $D_{\text{idk}}^1 = \text{sort}(D_{\text{idk}}^1, \text{key}=\sigma, \text{order=ascend})$
\State $D_{\text{van}}^2 = \text{TopK}(D_{\text{van}}^1, N_{\text{van}})$
\State $D_{\text{idk}}^2 = \text{TopK}(D_{\text{idk}}^1, N_{\text{idk}})$
\For{$x_i$ in $D_{\text{van}}^2$}
    \State $x_i.a_{\text{rait}}$ = $x_i.a$
\EndFor
\For{$x_j$ in $D_{\text{idk}}^2$}
    \State $x_j.a_{\text{rait}}$ = ``I don't know"
\EndFor
\State $D_{\text{rait}} = D_{\text{van}}^2 \cup D_{\text{idk}}^2$
\State \textbf{return} $D_{\text{rait}}$
\end{algorithmic}
\end{algorithm}

\section{Experimental Setup}

\subsection{Dataset}
We evaluate two tasks: knowledge-oriented Multiple Choice Questions Answering (MCQA) and Open-ended Questions Answering (OEQA). For MCQA, the MMLU \cite{MMLU} test split serves as the training set, MMLU val as the In-Domain (ID) test set, and ARC-c \cite{ARC_C} test split as the Out-Of-Domain (OOD) test set. For OEQA, the TriviaQA \cite{triviaqa} train split is used for training, TriviaQA dev for the ID test set, and NQ \cite{nq} dev for the OOD test set. More details are in Appendix D.1.

\subsection{Metric}
In post RAIT evaluation of LLMs, each test sample is classified as correct, incorrect, or refused. We calculate accuracy (\(P_c\)), error (\(P_w\)), and refusal rates (\(P_r\)) to assess performance, highlighting the key question: How to identify the better-performing model?

\subsubsection{Shortcomings of existing refusal-aware metrics}

We conducted the in-depth analysis of existing refusal-aware metrics, identifying several design shortcomings (see Appendix C.1). We highlighted these shortcomings through constructed examples, as shown in Table 1.

\begingroup
\fontsize{9}{10}\selectfont
\setlength{\tabcolsep}{1mm}
\begin{table}[h!]
\centering
{\fontsize{9pt}{10pt}\selectfont
\begin{tabular}{c|cc|cc}
\hline
Metric & $\mathcal{M}_1$ & $\mathcal{M}_2$ & $\mathcal{M}_3$ & $\mathcal{M}_4$ \\ \hline
$P_c\uparrow$ & 0.3 & 0.3 & 0.5 & 1 \\ 
$P_w\downarrow$ & 0.2 & 0.15 & 0 & 0 \\ 
$P_r$ & 0.5 & 0.55 & 0.5 & 0 \\ \hline
$S_{\text{honesty}}$ \cite{alignment_for_honesty} $\uparrow$ & (0.8) & (0.794) & (1) & (1) \\ 
TRUTHFUL \cite{cheng2024can} $\uparrow$ & (0.8) & (0.75) & 1 & 1 \\ 
rely \cite{rejection_improves} $\uparrow$ & (0.55) & (0.548) & 0.75 & 1 \\ 
R-Acc \cite{dont_hallucinate_abstain} $\uparrow$ & (0.8) & (0.778) & 1 & 1 \\ 
ER \cite{dont_hallucinate_abstain} $\uparrow$ & (0.3) & (0.25) & 0.5 & 1 \\ 
A-Acc  \cite{dont_hallucinate_abstain} $\uparrow$ & (0.8) & (0.75) & 1 & 1 \\ 
A-F1  \cite{dont_hallucinate_abstain} $\uparrow$ & (0.8) & (0.762) & 1 & 1 \\ 
AP \cite{R_Tuning} $\uparrow$ & — & — & (1) & (1) \\ \hline
THS (ours) $\uparrow$ & 0.1 & 0.15 & 0.5 & 1 \\ \hline
\end{tabular}
}
\caption{Comparison of refusal-aware metrics: The performance of constructed LLMs is \(\mathcal{M}_1 < \mathcal{M}_2 < \mathcal{M}_3 < \mathcal{M}_4\). However, existing metrics exhibit significant issues, as indicated by the numbers in (parentheses).}
\end{table}
\endgroup  
We constructed an initial model \(\mathcal{M}_0\) and four refined models \(\mathcal{M}_1\) to \(\mathcal{M}_4\), showing progressive improvement: \(\mathcal{M}_1 < \mathcal{M}_2 < \mathcal{M}_3 < \mathcal{M}_4\). Details on these models are in Appendix C.2. However, existing metrics have notable flaws: \textbf{\(S_{\text{honesty}}\)} \cite{alignment_for_honesty} ranks \(\mathcal{M}_1\) higher than \(\mathcal{M}_2\) and treats \(\mathcal{M}_3\) the same as \(\mathcal{M}_4\); \textbf{TRUTHFUL} \cite{cheng2024can} favors \(\mathcal{M}_1\) over \(\mathcal{M}_2\); and \textbf{R-Acc}, \textbf{ER}, \textbf{A-Acc}, and \textbf{A-F1} \cite{dont_hallucinate_abstain} also rank \(\mathcal{M}_1\) higher than \(\mathcal{M}_2\). Additionally, \textbf{AP} \cite{R_Tuning} fails to distinguish between \(\mathcal{M}_3\) and \(\mathcal{M}_4\).

\begin{figure}[h]
    \centering
    \vspace{-5pt}
    \includegraphics[width=2in]{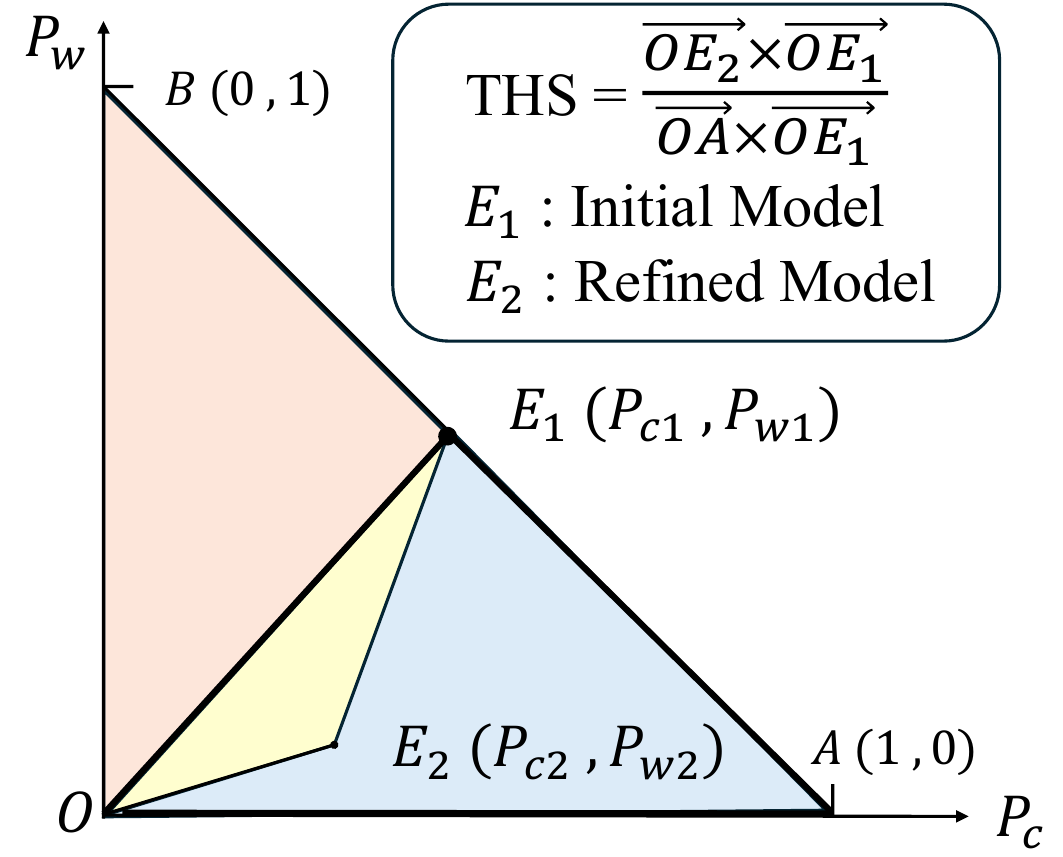}
    \caption{Truthful Helpfulness Score (THS).}
    \vspace{-5pt}
\end{figure}

\subsubsection{Our Metric: Truthful Helpfulness Score (THS)}

Due to the shortcomings of existing metrics, we propose the Truthful Helpfulness Score (THS).
We first establish a Cartesian coordinate system with $P_c$ and $P_w$ as axes, where point \(E_1\) represents the coordinates of the initial LLM, and point \(E_2\) represents the coordinates of the refined.
When \(E_2\) falls below \(OE_1\), a larger area of triangle \( \triangle OE_1E_2 \)
 indicates a stronger model. If \(E_2\) is above \(OE_1\), it suggests a decline in the model's performance. Based on this, we define THS as the ratio of the cross product of \(OE_1\) and \(OE_2\) to the maximum cross product value:
\[
\text{THS} = (\overrightarrow{OE_2} \times \overrightarrow{OE_1}) / (\overrightarrow{OA} \times \overrightarrow{OE_1})
\]

The results in Table 1 clearly demonstrate the effectiveness of THS. For a more detailed analysis of THS's effectiveness, please refer to Appendix C.3.

\subsection{Baselines}
To verify CRaFT's effectiveness, we compared it with mainstream methods:
\textbf{Init-Basic}: Uses the initial LLM with common question-answering prompts.  
\textbf{Init-Refuse}: Adds instructions like ``If you don't know, respond with `I don't know.'".  
\textbf{Van-Tuning}: Randomly selects \(N_{\text{van}} + N_{\text{idk}}\) samples from \(D_{\text{src}}\) for instruct-tuning without modification.  
\textbf{Cor-RAIT}: Implements the method from \cite{R_Tuning}, filtering and modifying RAIT data based on response correctness.  
Detailed prompts for each baseline are in Appendix D.2.

\subsection{Implementation Details}
In the experiments, we used LLaMA2-7B-Chat \cite{touvron2023llama} and LLaMA3-8B-Instruct \cite{dubey2024llama} as the initial LLM \(\mathcal{M}_0\). For the MCQA task, we selected 5000 samples from MMLU, and for the OEQA task, we used 10,000 samples from TriviaQA as training data. In all RAIT settings, except Van-Tuning, the ratio of vanilla to IdK samples was 1:4. We applied 5-shot and 3-shot knowledge state queries for the MCQA and OEQA tasks, respectively. Details on knowledge state and flow queries are in Appendix B.
For Instruct Tuning, we used XTuner~\footnote{https://github.com/InternLM/xtuner} with 3 epochs and a maximum context length of 2048. In MCQA, we applied LoRA \cite{hu2021lora} with settings \(r=64\), \(\alpha=16\), dropout=0.1, and a learning rate of 2e-4; for OEQA, full parameter training was used. More details on training are in Appendix D.3.
We used 0-shot and greedy decoding for evaluation, with further details in Appendix D.4. OpenCompass~\footnote{https://github.com/open-compass/opencompass} was employed for knowledge state queries and evaluations. All experiments were run on NVIDIA A100-80GB GPUs.

\section{Experimental Results and Analyses}

\begingroup
\fontsize{9}{11}\selectfont
\setlength{\tabcolsep}{1mm}
\renewcommand{\arraystretch}{1.2} 
\label{table:main table}
\begin{table*}[!t]
\centering
\resizebox{1.0\linewidth}{!}{
\begin{tabular}{ccc|ccc|ccc|ccc|ccc}
\hline
\multirow{3}{*}{\textbf{LLMs}} & \multicolumn{2}{c|}{\textbf{QA Type}} & \multicolumn{6}{c|}{\textbf{MCQA}} & \multicolumn{6}{c}{\textbf{OEQA}} \\
\cline{2-15}
& \multicolumn{2}{c|}{\textbf{Dataset}} & \multicolumn{3}{c|}{\textbf{MMLU (ID)}} & \multicolumn{3}{c|}{\textbf{ARC-c (OOD)}} & \multicolumn{3}{c|}{\textbf{TriviaQA (ID)}} & \multicolumn{3}{c}{\textbf{NQ (OOD)}} \\
\cline{2-15} 
& \multicolumn{2}{c|}{\textbf{Metric}} & $P_c$ & $P_w\downarrow$ & THS$\uparrow$ & $P_c$ & $P_w\downarrow$ & THS$\uparrow$ & $P_c$ & $P_w\downarrow$ & THS$\uparrow$ & $P_c$ & $P_w\downarrow$ & THS$\uparrow$ \\
\hline
\multirow{9}{*}{\shortstack{\textbf{LLaMA2-7B} \\ \textbf{Chat}}}
& \multirow{5}{*}{\textbf{Baselines}}  & Init-Basic &  45.6 & 52.8 & 00.0 & 53.9 & 46.0 & 00.0 & 54.0 & 46.0 & 00.0 & 29.3 & 70.7 & 00.0 \\ 
&& Init-Refuse & 36.4 & 38.9 & 03.9 & 44.4 & 35.7 & 02.6 & 37.0 & 21.7 & 11.5 & 20.8 & 38.6 & 04.8 \\ 
&& Van-Tuning & 46.9 & 53.1 & 01.2 & 54.5 & 45.5 & 01.2 & 48.6 & 44.5 & -03.7 & 18.3 & 50.2 & -02.5 \\ 
&& Cor-RAIT & 44.5 & 39.6 & 11.3 & 55.8 & 38.1 & 11.1 & 41.3 & 18.3 & 19.7 & 16.2 & 27.6 & 04.7 \\ \cdashline{2-15}
&\textbf{Ours}& CRaFT & 43.9 & 36.4 & 12.5 & 54.7 & 35.9 & 12.6 & 38.5 & \textbf{12.9} & \textbf{23.3} & 15.8 & 22.4 & \textbf{06.5} \\  \cdashline{2-15}
&\multirow{2}{*}{\textbf{Ablations}}& w/o Flow & 39.7 & \textbf{31.0} & \textbf{13.0} & 51.4 & \textbf{32.3} & 13.5 & 45.2 & 20.5 & 21.1 & 21.2 & 38.8 & 05.2 \\  
&& w/o Cer & 38.4 & 32.1 & 11.5 & 52.5 & 32.9 & \textbf{13.9} & 38.5 & 15.7 & 20.1 & 14.6 & \textbf{22.1} & 05.4 \\  
\hline
\multirow{9}{*}{\shortstack{\textbf{LLaMA3-8B} \\ \textbf{Instruct}}}
& \multirow{5}{*}{\textbf{Baselines}} & Init-Basic & 66.8 & 33.1 & 00.0 & 80.6 & 19.5 & 00.0 & 66.8 & 33.2 & 00.0 & 40.3 & 59.7 & 00.0 \\ 
&& Init-Refuse & 50.0 & 17.0 & 15.6 & 65.3 & 14.4 & 05.6 & 53.9 & 20.8 & 12.0 & 31.1 & 38.6 & 05.0 \\ 
&& Van-Tuning & 69.5 & 30.5 & 08.0 & 80.3 & 19.7 & -01.3 & 55.0 & 38.1 & -21.8 & 21.0 & 48.5 & -11.7 \\ 
&& Cor-RAIT & 63.9 & 21.6 & 20.4 & 79.4 & 16.2 & 12.2 & 45.4 & 13.2 & 18.8 & 17.2 & 25.6 & -00.1 \\ \cdashline{2-15}
&\textbf{Ours}& CRaFT & 53.3 & \textbf{09.6} & \textbf{34.0} & 74.1 & \textbf{12.7} & \textbf{21.4} & 43.5 & \textbf{10.9} & \textbf{21.5} & \textbf{19.0} & 27.5 & \textbf{00.4} \\  \cdashline{2-15}
&\multirow{2}{*}{\textbf{Ablations}}& w/o Flow & 57.5 &15.3 & 27.2 & 75.8 & 14.9 & 13.9 & 49.1 & 18.0 & 12.8 & 22.3 & 41.6 & -05.8 \\  
&& w/o Cer & 62.1 & 18.4 & 25.0 & 78.2 & 17.3 & 06.5 & 43.0 & 11.2 & 20.5 & 15.8 & \textbf{23.5} & -00.1 \\  
\hline
\end{tabular}}
\caption{Performance comparisons on MMLU, ARC-c, TriviaQA and NQ. The best performance is highlighted in \textbf{boldface}.}
\vspace{-10pt}
\end{table*}
\endgroup

\subsection{Overall Performance}
The experimental results on the OEQA and MCQA tasks are presented in Table 2. Under the ID setting for both types of tasks, our method outperformed four baseline models on THS, achieving the best results. Specifically, under the ID setting for OEQA, compared to the current best RAIT baseline, CRaFT improved the THS on LLaMA2-7B-Chat and LLaMA3-8B-Instruct by 3.56 and 2.72, respectively. Similarly, under the ID setting for MCQA, CRaFT improved the THS by 1.14 and 13.57, respectively. This indicates that CRaFT can significantly improve the model's rejection capability under the ID setting. Under the OOD setting, CRaFT improved the THS on the MCQA task by 1.5 and 9.2, respectively, compared to Cor-RAIT. On the OEQA's LLaMA2-7B-Chat, it improved by 1.76 compared to the most competitive method, Init-Refuse. Overall, CRaFT demonstrated excellent competitiveness in model generalization. Furthermore, we found that on the MCQA task, compared to other baselines, Cor-RAIT showed significant improvements under both ID and OOD settings. However, on the OEQA task, Cor-RAIT performed worse than Init-Refuse under the OOD setting. This reveals the limitations of the instruction fine-tuning method. It's worth mentioning that Van-Tuning generally had a negative impact on the improvement of overall capability, implying that the instruction fine-tuning approach of forcing the model to answer can undermine the model's inherent rejection capability. Therefore, although CRaFT surpassed Cor-RAIT under all tasks and settings, the improvement was limited under the OOD setting for OEQA due to training paradigm.

\subsection{Ablation Experiments}
In order to resolve the static and dynamic conflicts that lead to over-refusal, we extend Cor-RAIT to construct RAIT data using the information of correctness, certainty, and knowledge flow. We conduct sufficient ablation experiments to deeply investigate the impact of the above three factors on RAIT data selection. Compared to Cor-RAIT, the method only introducing response certainty which named as ``w/o Flow" achieved significant gains on the THS in the MCQA and OEQA tasks. This indicates that eliminating static conflicts can effectively mitigate the over-refusal of LLMs and this improvement is generalizable. ``w/o Cer" only uses response correctness and knowledge flow. Similarly, experimental results show that introducing knowledge flow to filter dynamic conflicts can also maintain the factuality of the model while improving its rejection capability. Finally, CRaFT considers both static and dynamic conflicts, further enhancing performance improvement.

\section{Conclusion}

In this paper, we identify over-refusal in correctness-based RAIT methods, caused by static and dynamic conflicts in RAIT data. To address this, we propose CRaFT: it mitigates static conflicts by incorporating response certainty during data construction and overcomes dynamic conflicts through rehearsal training to capture knowledge flow trends in LLMs. Extensive experiments on MCQA and OEQA tasks show CRaFT outperforms existing baselines, validating its effectiveness. Future work includes enhancing CRaFT with RL-based strategies and adapting it for more complex tasks, such as reasoning and multi-turn dialogue.

\section*{Acknowledgments}
This research was supported by Shanghai Artificial Intelligence Laboratory. 

\bibliography{ref}
\appendix
\section{Static and Dynamic Conflicts}

\subsection{More t-SNE visualization of the LLM feature space}
Building on our previous analysis, where we visualized the latent representations of MMLU test samples using the LLaMA3-8B-Instruct model, we extended the t-SNE analysis to include both the MMLU and TriviaQA datasets across different models, specifically LLaMA2-7B-Chat and LLaMA3-8B-Instruct. The t-SNE plots in Figure 8 highlight significant overlaps between IdK and Vanilla samples within Cor-RAIT, consistently observed across models and datasets, reinforcing our findings on representation spaces. Meanwhile, CorCer-RAIT demonstrates clearer separations, effectively reducing static conflict compared to Cor-RAIT.

\begin{figure*}[t]
    \centering
    \vspace{-5pt}
    \includegraphics[width=6.0in]{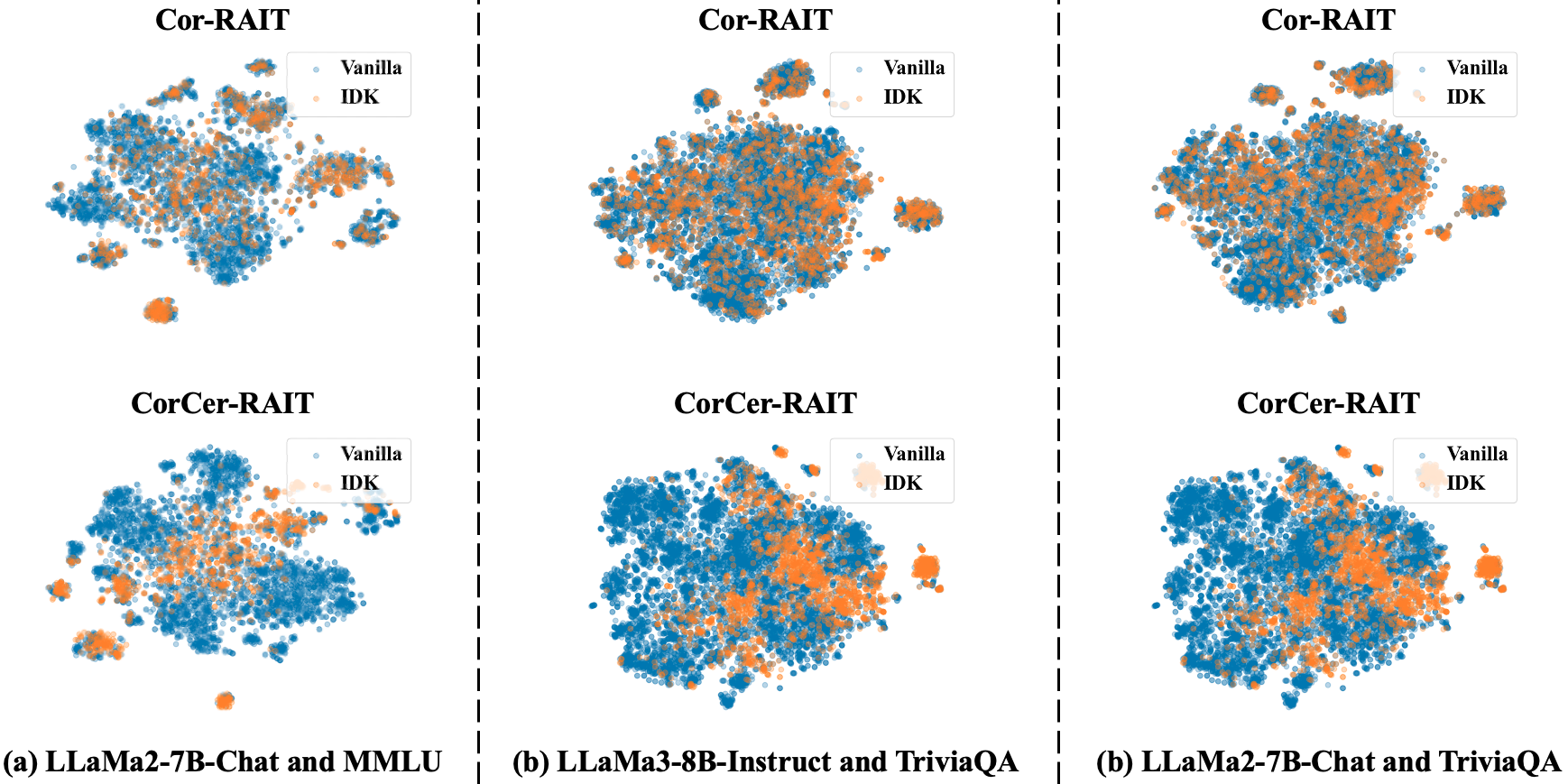}
     \caption{More t-SNE visualization of the LLM feature space}
    \label{fig:appendix_t-sne}
\end{figure*}

\subsection{Theoretical Analysis of the Relation between correctness and LLM representation}
\label{A2}
Through quantitative experimental analysis, we observed a relatively weak correlation between the latent representation and the correctness of predictions. Even small changes in the representation can lead to significant differences in correctness. We aim to delve deeper into the reasons behind this phenomenon.

We define the representation (i.e., the hidden state of the last token) as \(r \in \mathbb{R}^d\), where \(r\) is the dimension of the hidden layer. The linear layer's parameters include a weight matrix \(W \in \mathbb{R}^{V \times d}\) and a bias vector \(b \in \mathbb{R}^V\). The logits output is given by:
\[
z = W r + b
\]

The softmax function converts the logits into a probability distribution \(p\), where the probability of the \(i\)-th token is:
\[
p_i = \frac{e^{z_i}}{\sum_{j=1}^V e^{z_j}} = \frac{e^{W_i r + b_i}}{\sum_{j=1}^V e^{W_j r + b_j}}, \quad i = 1, 2, \ldots, V
\]

The gradient of the hidden layer representation \(r\) with respect to the probability of the ground truth token \(p_t\) is:
\[
\frac{\partial p_t}{\partial r} = p_t \left( W_t - \sum_{j=1}^V p_j W_j \right)
\]

This shows that the gradient depends on the difference between the correct token's weight \(W_t\) and the weighted average of all token weights, leading to larger adjustments in \(r\) when this difference is significant. While the denominator in the softmax is continuous, the numerator can cause discontinuities in \(p_t\) because small changes in \(r\) can abruptly alter the predicted token, leading to sudden drops in \(p_t\). Even small changes in \(r\) can have a significant impact on \(p_t\) due to the inherent discontinuities, often leading to highly similar samples being classified into \(D_{\text{van}}\) and \(D_{\text{idk}}\), respectively.

\subsection{Theoretical Analysis of the Relation between Certainty and LLM Representation}
\label{A3}
Entropy serves as a measure of certainty, indicating the model's confidence in its predictions. The entropy of the probability distribution \( p \) is defined as:
\[
H = -\sum_{i=1}^V p_i \log p_i
\]

The gradient of entropy with respect to the hidden layer representation \( r \) is:
\[
\frac{\partial H}{\partial r} = - \sum_{i=1}^V (\log p_i + 1) p_i (W_i - p_i W)
\]

For small changes in the representation \( dr \), the change in entropy can be approximated using a second-order expansion:
\[
dH \approx \nabla H(r)^T dr + \frac{1}{2} dr^T \nabla^2 H(r) dr
\]

Since both the entropy function \( H(p) \) and its gradient are continuous, small changes in representation \( dr \) result in minimal changes in entropy \( dH \). This aligns with our observation that minor variations in representation generally have a limited impact on certainty. Accordingly, we can ensure differences in certainty between \(D_{\text{van}}\) and \(D_{\text{idk}}\) to maintain distinctions in their representations, thereby mitigating static conflict.

\subsection{Statistics of Static Conflicts}
In addition to our previous analysis, we applied CRSS to the TriviaQA dataset, as shown in Figure 9. The results indicate that introducing certainty to some extent decreases the rate of conflicting samples in open-domain question answering. This reduction suggests that our approach effectively mitigates the negative impact of conflicting supervision signals, thereby reducing the likelihood of over-refusal.

\begin{figure}[th]
    \centering
    \vspace{-5pt}
    \includegraphics[width=3.3in]{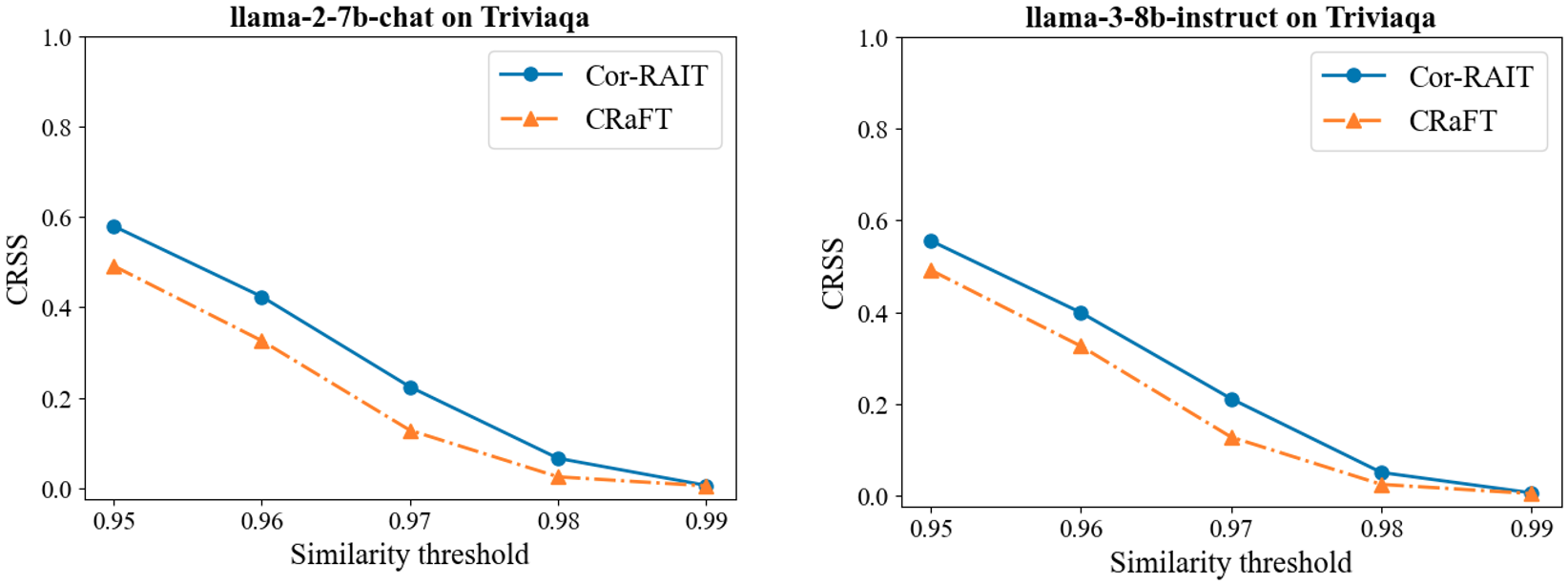}
    \caption{CRSS of Different RAIT Samples on TriviaQA.}
    \vspace{-5pt}
\end{figure}

\subsection{Statistics of Dynamic Conflicts}
\label{A5}
\begingroup
\begin{table}[h!]
\setlength{\tabcolsep}{1mm}
\centering
\begin{tabular}{|c|c|c|c|c|}
\hline
\textbf{Dataset} & \textbf{$n_{\text{rhs}}$} & \textbf{$n_{\mu < 0.5}$} & \textbf{$n_{\Delta \mu > 0}$} & \textbf{$n_{\mu < 0.5,\Delta \mu > 0}$} \\ \hline
\multirow{4}{*}{\textbf{MMLU}} & \textbf{200}   & 5175  & 4812  & 3318   \\ \cline{2-5}
                      & \textbf{500}   & 5175  & 5308  & 3252   \\ \cline{2-5}
                      & \textbf{1000} & 5175  & 4740  & 3571   \\ \cline{2-5}
                      & \textbf{1629}  & 5175  & 4389  & 3551   \\ \hline
\multirow{4}{*}{\textbf{TriviaQA}} & \textbf{200}   & 42457 & 9188  & 5102   \\ \cline{2-5}
                          & \textbf{1000} & 42457 & 9188  & 5034   \\ \cline{2-5}
                          & \textbf{3000} & 42457 & 9219  & 5061   \\ \cline{2-5}
                          & \textbf{8000} & 42457 & 11302 & 9553   \\ \hline
\end{tabular}
\caption{Comparison of Dataset Statistics across Different $n_{rhs}$ Values.}
\end{table}
\endgroup  

To capture the natural knowledge flow of the LLM during SFT, we conducted experiments using rehearsal training. We tested the LLaMA3-8B-Instruct on the MMLU dataset (14,042 samples) and the TriviaQA dataset (87,622 samples). The amount of training data for rehearsal training, denoted as $n_{\text{rhs}}$, was varied. For MMLU, $n_{\text{rhs}}$ values were set to 200, 500, 1000, and 1629, with 1629 representing the full set of training data filtered by correctness greater than 0.99. For TriviaQA, $n_{\text{rhs}}$ values were 200, 1000, 3000, and 8000.
In the experiment, we denote the accuracy of the initial model on a given sample as \(\mu\), the accuracy of the model after rehearsal training as \(\tilde{\mu}\), and the difference between the two as \(\Delta \mu\). Key metrics for the samples are summarized in Table 3. Regardless of the value of \(n_{\text{rhs}}\), we found that a significant portion of samples with initial correctness less than 0.5 experienced an increase in \(\mu\).

\section{Details of CRaFT}

\begin{table*}[h!]
\begin{tabular}{|c|l|}
\hline
\centering
\textbf{Query Type}                                                                & \multicolumn{1}{c|}{\textbf{Instruction}}                                                                                                                                                                                                                                                                                                                                                                      \\ \hline
\begin{tabular}[c]{@{}c@{}}\textbf{Knowledge State Query} \\ \textbf{based on} \\ \textbf{TOKEN Probability}\end{tabular}    & \begin{tabular}[c]{@{}l@{}}There is a single choice question about \{\textit{Task}\}. Answer the question by replying A, B, C or D.\\ 
\textbf{Question}: \{\textit{Question}\} \\
\textbf{A}. \{\textit{Context\_of\_A}\} \\
\textbf{B}. \{\textit{Context\_of\_B}\} \\
\textbf{C}. \{\textit{Context\_of\_C}\} \\
\textbf{D}. \{\textit{Context\_of\_D}\} \\
\textbf{Answer}:\textbf{\{\textit{Target\_Token}\}}(The Target Token takes on the values A, B, C, and D.)\end{tabular}                                                                                                                                                                                                                                                                                                    \\ \hline
\begin{tabular}[c]{@{}c@{}}\textbf{Knowledge State Query} \\ \textbf{based on} \\ \textbf{Generation}\end{tabular} & \begin{tabular}[c]{@{}l@{}}There is a single choice question about \{\textit{Task}\}. Answer the question by replying A, B, C or D.\\
\textbf{Question}: \{\textit{Question}\} \\
\textbf{A}. \{\textit{Context\_of\_A}\} \\
\textbf{B}. \{\textit{Context\_of\_B}\} \\
\textbf{C}. \{\textit{Context\_of\_C}\} \\
\textbf{D}. \{\textit{Context\_of\_D}\} \\
\textbf{Answer}:\end{tabular}                                                                                                                        \\ \hline                                                                                                                                                                                                                                                                                                                                         
\end{tabular}
\label{table:knowledege_state_query}
\caption{Prompt templates of the two type knowledge state query. The italic \{\textit{text}\} in curly braces represents variables that need to be replaced.}
\end{table*}

\subsection{Two Types of Knowledge State Query}
Our research focuses on the MCQA and OEQA tasks, which correspond to token probability-based knowledge state queries and generation-based knowledge state queries, respectively. In the MCQA task, we use an in-context learning approach with 5-shot examples to help the model understand the task. The instruction format is shown in Table 4, where \textbf{\textit{Target\_Token}} takes on the values A, B, C, and D, allowing us to calculate the corresponding token probabilities for correctness and certainty. In the OEQA task, we use a 3-shot setup. The instruction format, as shown in Table 5, involves setting the temperature parameter to 1 and enabling \texttt{do\_sample} for $N$ iterations of inference, followed by calculations of correctness and certainty.

\subsection{Rehearsal Training}
\label{B2}

\subsubsection{Details about rehearsal training}

For both of the OEQA and MCQA, the number of training data samples are 1000. 
We train set for rehearsal training from $D_{\text{src}}$ according to the knowledge state $\mathcal{M}_{0}$. 
For OEQA task, the training data $D_{\text{rehearsal}} = \{x_i|x_i.\mu\ge0.995~\text{ and }~x_i.\tau\ge0.995\}$.
For MCQA task, the training data $D_{\text{rehearsal}} = \{x_i|x_i.\mu\ge0.99 \}$.

In the MCQA tasks, we used LoRA, setting $r=64$, $\alpha=16$, $dropout=0.1$; in the OEQA tasks, we used full parameter training.
The training is 3 epochs, using the cosine learning rate scheduler \textbf{without} warming up. 
The maximum learning rate for MCQA task is 2e-4. The maximum learning rate on OEQA for LLaMA2-7B-Chat is 2e-5, for LLaMA3-8B-Instruct is 1e-6. 

\begingroup
\fontsize{9}{11}\selectfont
\setlength{\tabcolsep}{1mm}
\begin{table}[h!]
\centering
\begin{tabular}{|c|c|lll|lll|}
\hline
\multicolumn{1}{|l|}{}          & \textbf{}             & \multicolumn{3}{c|}{\textbf{MMLU (ID)}}                                                                       & \multicolumn{3}{c|}{\textbf{ARC-c (OOD)}}                                                                     \\ \hline
\textbf{Method}                 & \textbf{$n_{\text{rhs}}$} & \multicolumn{1}{c|}{\textbf{$P_c$}$\uparrow$} & \multicolumn{1}{c|}{\textbf{$P_w$}$\downarrow$} & \multicolumn{1}{c|}{\textbf{THS}$\uparrow$} & \multicolumn{1}{c|}{\textbf{$P_c$}$\uparrow$} & \multicolumn{1}{c|}{\textbf{$P_w$}$\downarrow$} & \multicolumn{1}{c|}{\textbf{THS}$\uparrow$} \\ \hline
\textbf{Init-Basic}             & \textbf{-}            & \multicolumn{1}{l|}{66.9}           & \multicolumn{1}{l|}{33.1}           & 0                                 & \multicolumn{1}{l|}{80.6}           & \multicolumn{1}{l|}{19.5}           & 0                                 \\ \hline
\textbf{Init-Refuse}            & \textbf{-}            & \multicolumn{1}{l|}{50}             & \multicolumn{1}{l|}{17}             & 15.6                              & \multicolumn{1}{l|}{65.3}           & \multicolumn{1}{l|}{14.4}           & 5.6                               \\ \hline
\textbf{Van-Tuning}             & \textbf{-}            & \multicolumn{1}{l|}{69.5}           & \multicolumn{1}{l|}{30.5}           & 8                                 & \multicolumn{1}{l|}{80.3}           & \multicolumn{1}{l|}{19.7}           & -1.3                              \\ \hline
\textbf{Cor-RAIT}               & \textbf{-}            & \multicolumn{1}{l|}{63.9}           & \multicolumn{1}{l|}{21.6}           & 20.4                              & \multicolumn{1}{l|}{66.8}           & \multicolumn{1}{l|}{16.2}           & 12.2                              \\ \hline
\multirow{4}{*}{\textbf{CRaFT}} & \textbf{200}          & \multicolumn{1}{l|}{54.1}           & \multicolumn{1}{l|}{12.2}           & 27.7                              & \multicolumn{1}{l|}{74.5}           & \multicolumn{1}{l|}{14.2}           & 15.9                              \\ \cline{2-8} 
                                & \textbf{500}          & \multicolumn{1}{l|}{55.8}           & \multicolumn{1}{l|}{13.6}           & 28.4                              & \multicolumn{1}{l|}{76}             & \multicolumn{1}{l|}{15}             & 13.8                              \\ \cline{2-8} 
                                & \textbf{1000}         & \multicolumn{1}{l|}{53.3}           & \multicolumn{1}{l|}{9.6}            & 34                                & \multicolumn{1}{l|}{74.1}           & \multicolumn{1}{l|}{12.7}           & 21.4                              \\ \cline{2-8} 
                                & \textbf{1629}    & \multicolumn{1}{l|}{54.7}           & \multicolumn{1}{l|}{11.4}           & 31.2                              & \multicolumn{1}{l|}{75}             & \multicolumn{1}{l|}{14.8}           & 13.9                              \\ \hline
\end{tabular}
{\fontsize{9pt}{11pt}\selectfont
\caption{Results of Reheasal Training in MCQA Task ($\mathcal{M}_{0}$ is LLaMA3-8B-Instruct).}
\label{table:ab_MCQA_rehearsal}}
\end{table}
\endgroup  
\begingroup
\fontsize{9}{11}\selectfont
\setlength{\tabcolsep}{1mm}
\begin{table}[h!]
\centering
\begin{tabular}{|c|c|lll|lll|}
\hline
\multicolumn{1}{|l|}{}          & \textbf{}             & \multicolumn{3}{c|}{\textbf{TriviaQA (ID)}}                                                                   & \multicolumn{3}{c|}{\textbf{NQ (OOD)}}                                                                        \\ \hline
\textbf{Method}                 & \textbf{$n_{\text{rhs}}$} & \multicolumn{1}{c|}{\textbf{$P_c$}$\uparrow$} & \multicolumn{1}{c|}{\textbf{$P_w$}$\downarrow$} & \multicolumn{1}{c|}{\textbf{THS}$\uparrow$} & \multicolumn{1}{c|}{\textbf{$P_c$}$\uparrow$} & \multicolumn{1}{c|}{\textbf{$P_w$}$\downarrow$} & \multicolumn{1}{c|}{\textbf{THS}$\uparrow$} \\ \hline
\textbf{Init-Basic}             & \textbf{-}            & \multicolumn{1}{l|}{54}             & \multicolumn{1}{l|}{46}             & 0                                 & \multicolumn{1}{l|}{29.3}           & \multicolumn{1}{l|}{70.7}           & 0                                 \\ \hline
\textbf{Init-Refuse}            & \textbf{-}            & \multicolumn{1}{l|}{37}             & \multicolumn{1}{l|}{21.8}           & 11.5                              & \multicolumn{1}{l|}{20.8}           & \multicolumn{1}{l|}{38.6}           & 4.8                               \\ \hline
\textbf{Van-Tuning}             & \textbf{-}            & \multicolumn{1}{l|}{48.6}           & \multicolumn{1}{l|}{44.5}           & -3.7                              & \multicolumn{1}{l|}{18.3}           & \multicolumn{1}{l|}{50.2}           & -2.5                              \\ \hline
\textbf{Cor-RAIT}               & \textbf{-}            & \multicolumn{1}{l|}{41.3}           & \multicolumn{1}{l|}{18.3}           & 19.7                              & \multicolumn{1}{l|}{16.2}           & \multicolumn{1}{l|}{27.6}           & 4.7                               \\ \hline
\multirow{4}{*}{\textbf{CRaFT}} & \textbf{200}          & \multicolumn{1}{l|}{39}             & \multicolumn{1}{l|}{13.5}           & 23.2                              & \multicolumn{1}{l|}{16.5}           & \multicolumn{1}{l|}{24.9}           & 6.2                               \\ \cline{2-8} 
                                & \textbf{1000}         & \multicolumn{1}{l|}{38.5}           & \multicolumn{1}{l|}{12.9}           & 23.3                              & \multicolumn{1}{l|}{15.8}           & \multicolumn{1}{l|}{22.4}           & 6.5                               \\ \cline{2-8} 
                                & \textbf{3000}         & \multicolumn{1}{l|}{41.1}           & \multicolumn{1}{l|}{14.7}           & 23.9                              & \multicolumn{1}{l|}{18.1}           & \multicolumn{1}{l|}{28.2}           & 6.4                               \\ \cline{2-8} 
                                & \textbf{8000}         & \multicolumn{1}{l|}{40.2}           & \multicolumn{1}{l|}{14.4}           & 23.3                              & \multicolumn{1}{l|}{17}             & \multicolumn{1}{l|}{25}             & 6.7                               \\ \hline
\end{tabular}
{\fontsize{9pt}{11pt}\selectfont
\caption{Results of Reheasal Training in OEQA Task. ($\mathcal{M}_{0}$ is LLaMA2-7B-Chat)}
\label{table:ab_OEQA_rehearsal}}
\end{table}
\endgroup  

\subsubsection{Ablation experiments of rehearsal training}

In our work, a rehearsal training mechanism is proposed for detecting the knowledge flow generated by LLM during the SFT process. 
To study the impact of different levels of training on knowledge flow detection and the final experimental results of CRaFT, we adjusted the sample size of the rehearsal training set, \( n_{\text{rhs}} \). Starting from \( n_{\text{rhs}} = 1000 \), for MCQA, we set \( n_{\text{rhs}} \in \{200, 500, 1000, 1629\footnote{The number of all the samples that meet the 'rehearsal training' requirements is 1629 for MCQA task.}\} \); for OEQA, we set \( n_{\text{rhs}} \in \{200, 1000, 3000, 8000\} \). Hyperparameters other than \( n_{\text{rhs}}\) remained constant. The experimental results are shown in Table 5 and Table 6.

 The results show that, compared to other methods, regardless of the size of the rehearsal training dataset, CRaFT's THS is always the highest. This indicates that rehearsal training is not sensitive to the size of the training data and has robustness. In addition, we found that with the change in \(n_{rhs}\), CRaFT does not perform the same on OEQA and MCQA. On the OEQA task, CRaFT's THS is basically stable; while on MCQA, its THS fluctuates more noticeably. This may be due to the different difficulty levels of the datasets used in the two tasks. Objectively, the OEQA task dataset we chose is simpler than the MCQA task dataset.

\section{More Details about Metrics}
\subsection{Details and Shortcomings of existing refusal-aware metrics}
Previously, we mentioned that RAIT (Refusal-Aware Instruction Tuning) has two main effects on the model: it enhances refusal-aware capabilities and alters the model's internal knowledge state, impacting both honesty and helpfulness. A good metric should be both \textbf{singular} and \textbf{comprehensive}, meaning that we should be able to evaluate the quality of the model based on this single metric. It should effectively balance honesty and helpfulness without requiring additional hyperparameters.

Unfortunately, after conducting a systematic review of the metrics used in existing refusal-aware work, we found that none of the proposed metrics meet the above criteria in RAIT. These metrics often emphasize one specific aspect of the model's capabilities. In this section, we will conduct a detailed analysis of the existing metrics and highlight their shortcomings.

Alignment for Honesty \cite{alignment_for_honesty} proposes a method that observes changes in a model’s response types before (t) and after (t+1) alignment for honesty, which is the most granular analysis method in current research. We will also adopt this method in our subsequent analysis of existing approaches. The paper introduces a metric, $S_{\text{honesty}} = \frac{1}{2} (S_{\text{prudence}} + (1 - S_{\text{over-consv.}}))$, which seems to be a comprehensive metric. However, during our experiments, we found that Model A performed better than Model B in both correct and incorrect responses, yet its $S_{\text{honesty}}$ score was lower than Model B’s. This discrepancy arises because the metric does not fully account for changes in the model’s internal knowledge state, particularly when some samples shift from correct to wrong or from wrong to correct. We will provide specific examples to illustrate this issue in the following sections.

\cite{cheng2024can} introduces the TRUTHFUL RATE, which is the sum of the proportion of questions correctly answered by the model and the proportion of questions that the model correctly refuses to answer. The issue with using this method in RAIT is that it relies on the initial model to classify data into known and unknown categories, and then evaluates the refusal ability of the modified model. However, after RAIT, the model’s internal state changes, meaning the classification of known and unknown data should also be adjusted accordingly. Therefore, TRUTHFUL RATE is not suitable as a metric for RAIT.

\cite{dont_hallucinate_abstain} uses four metrics: Reliable Accuracy (R-Acc), Effective Reliability (ER), Abstain Accuracy (A-Acc), and Abstain F1 (A-F1). This clearly violates the principle of singularity. Additionally, like \cite{cheng2024can}, it fails to account for the conversion between correct and wrong samples, making these metrics inadequate.

\cite{rejection_improves} defines precision (\text{Prec}) as $ \frac{N_c}{N_c + N_w} $, where $ N_c $, $ N_r $, and $ N_w $ represent the counts of correct, rejected, and wrong responses, respectively. Accuracy (\text{Acc}) is defined as $ \frac{N_c}{N} $, and truthfulness (\text{Truth}) is given by $ 1 - \frac{N_w}{N} $. The answer rate (\text{Ans}) is $ 1 - \frac{N_r}{N} $, representing the proportion of non-rejected responses. The overall reliability (\text{Rely}) metric is then:
\[
\text{Rely} = \text{Ans} \cdot \text{Truth} + (1 - \text{Ans}) \cdot \text{Acc}.
\]

We further simplified \text{Rely} to $ \text{Rely} = \text{Prec} + \text{Ans} - (\text{Ans})^2 $, under the condition that $ \text{Prec} + \text{Ans} < 1 $. When $ \text{Prec} $ is fixed, $ \text{Ans} $ increases in the interval (0, 0.5) and decreases in the interval (0.5, 1). This shows that $ \text{Rely} $ does not necessarily increase as $ \text{Ans} $ increases, which is clearly problematic. This is clearly unreasonable, and I will provide examples in the following sections to demonstrate its issues.

R-Tuning \cite{R_Tuning} introduces the AP metric, which only considers the accuracy of willingly answered questions. This means the focus is primarily on honesty, without adequately considering helpfulness. For instance, if the model refuses to answer most questions and only responds to a few, but does so correctly, the AP value would still be high. Therefore, this metric is also unreasonable.

In summary, the existing metrics used in RAIT are inadequate because they fail to meet the criteria of being both singular and comprehensive. While some metrics focus on specific aspects like honesty or helpfulness, they often neglect the balance between the two or require multiple measurements to assess a model's performance. Additionally, these metrics do not fully account for changes in the model's internal knowledge state after training, which is crucial for accurately evaluating RAIT. A truly effective metric should provide a holistic evaluation of the model's refusal-aware capabilities without introducing additional complexity or hyperparameters.

\subsection{Details of \( \mathcal{M}_0 \) to \( \mathcal{M}_4 \)}
We adopt the method proposed in Alignment for Honesty \cite{alignment_for_honesty} to observe the state changes of each sample before and after model fine-tuning to evaluate the model. We assume a test set \(D_{\text{test}}\) with \(N=100\) samples, and define an initial model \( \mathcal{M}_0 \). For the initial model, 50 samples are answered correctly, and 50 samples are answered incorrectly, with corresponding \(P_c\) and \(P_w\) both being 0.5. We construct four refined models \( \mathcal{M}_1 \) to \( \mathcal{M}_4 \). The state changes of \( \mathcal{M}_1 \) and \( \mathcal{M}_2 \) are shown in Table 7. \( \mathcal{M}_3 \) represents the case where all previously wrong samples are changed to refuse after the model is refined, resulting in an accuracy of 0.5, a refusal rate of 0.5, and an error rate of 0. \( \mathcal{M}_4 \) is the case where all samples become correct, resulting in an accuracy of 1.

\begin{table}[h!]
\centering
\fontsize{9}{10}{
\begin{tabular}{c|ccc|ccc}
\hline
\multirow{2}{*}{\diagbox{\( \mathcal{M}_0 \)}{\( \mathcal{M}_{t} \)}} & \multicolumn{3}{c|}{\( \mathcal{M}_1 \)} & \multicolumn{3}{c}{\( \mathcal{M}_2 \)} \\
\cline{2-7}
 & $\mathbf{N_C}$ & $\mathbf{N_W}$ & $\mathbf{N_R}$ & $\mathbf{N_C}$ & $\mathbf{N_W}$ & $\mathbf{N_R}$ \\
\hline
$\mathbf{N_C}$ & 30 & 10 & 10 & 25 & 10 & 15 \\
$\mathbf{N_W}$ & 0  & 10 & 40 & 5  & 5  & 40 \\
$\mathbf{N_R}$ & 0  & 0  & 0  & 0  & 0  & 0  \\
\hline
\end{tabular}}
\caption{State Changes for the Refined LLMs, \( \mathcal{M}_1 \) and \( \mathcal{M}_2 \), from the Initial LLM \( \mathcal{M}_0 \)}
\label{tab:data}
\end{table}

\subsection{Effectiveness Analysis of THS}
We argue for the effectiveness of our metric from two perspectives: one is through a point-by-point analysis, and the other is by evaluating the impact of RAIT on the model.

When \(E_2\) falls at (0, 0), the THS is 0. When \(E_2\) falls at \((P_{c1}, 0)\), which corresponds to the model \( \mathcal{M}_3 \), it indicates that the model refuses to answer all incorrect questions, maximizing its refusal-aware capability. When \(E_2\) falls at (1, 0), the model achieves peak helpfulness by correctly answering all questions. Under the THS metric, the performance of these points increases sequentially.

Further analysis shows that when \(P_{c2}\) is fixed, the smaller \(P_{w2}\) is, the larger the area of triangle \(\triangle OE_1E_2\), resulting in a higher THS. Conversely, when \(P_{w2}\) is fixed, the smaller \(P_{c2}\) is, the smaller the area of triangle \(OE_1E_2\), and the lower the THS. This aligns well with the evaluation standards for model performance.

From another perspective, RAIT enhances refusal-aware capability, contributing to honesty, while also altering the model's knowledge state, impacting helpfulness. If we only consider changes in helpfulness, the impact of RAIT on the model is reflected in the vector \( \overrightarrow{E_1E_2} \) in Figure 10.a, which is parallel to vector \( \overrightarrow{AB} \). The closer \(E_2\) is to A, the stronger the model's helpfulness. If we only consider changes in refusal-aware capability, the variations in \(P_c\) and \(P_w\) together form the vector \( \overrightarrow{E_1E_2} \). If \(E_2\) lies below \(OE_1\), it indicates that a higher proportion of wrong samples have been converted to refusals, significantly enhancing the model's refusal-aware capability. However, if \(E_2\) falls on the line \(OE_1\), even with a high refusal rate, THS will not increase. If \(E_2\) is above \(OE_1\), it indicates a severe over-refusal phenomenon, leading to a negative THS. Based on the above analysis, THS comprehensively measures the model's honesty and helpfulness.

\begin{figure}[h]
    \centering
    \vspace{-5pt}
    \includegraphics[width=3.4in]{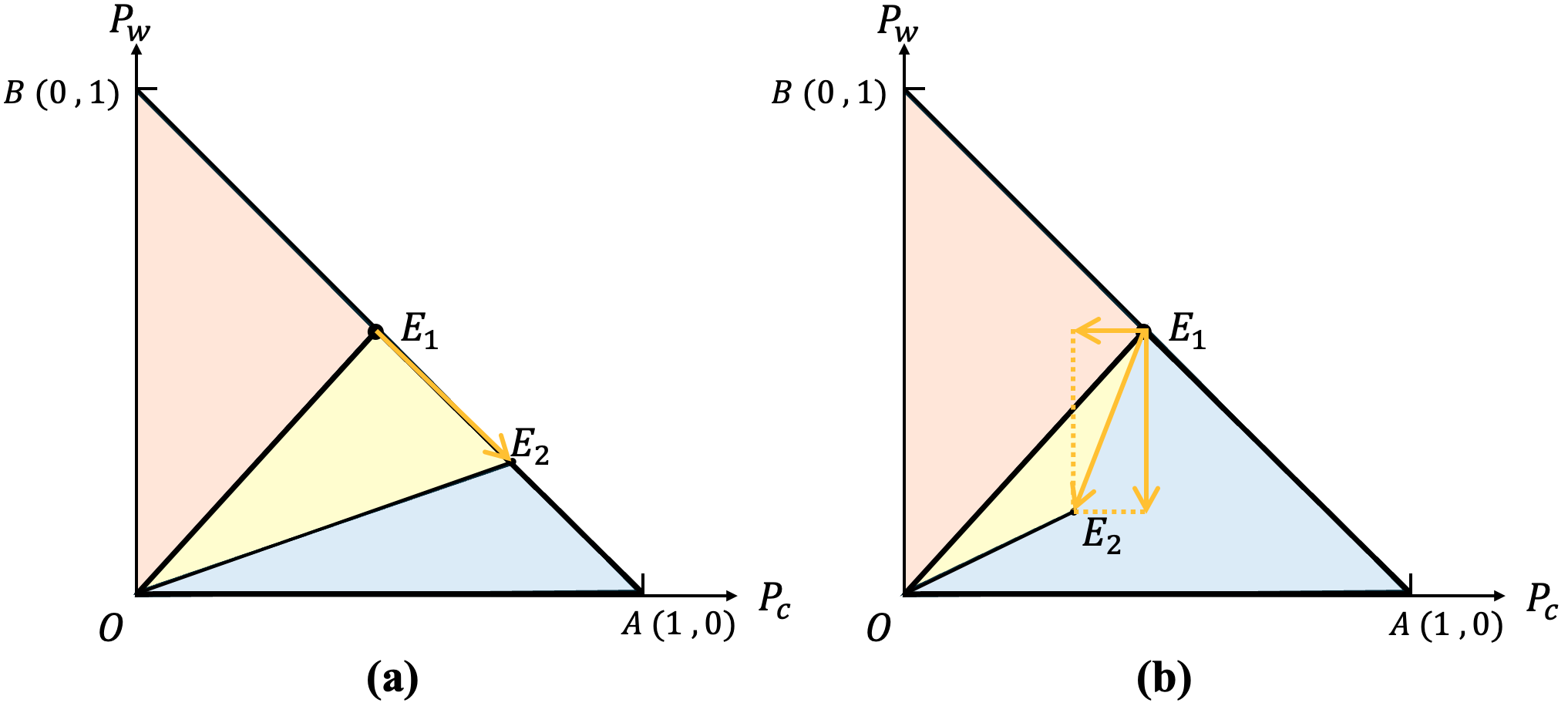}
    \caption{Analysis of THS.}
    \label{fig:THS}
    \vspace{-5pt}
\end{figure}

\section{More Details about Experiments}

\subsection{Datasets}
In this paper, we evaluate two tasks: knowledge-oriented Multiple Choice Questions Answer (MCQA) and Open-ended Questions Answer (OEQA). For the MCQA task, the MMLU \cite{MMLU} test split is used as the training dataset. The MMLU val split and the ARC-c \cite{ARC_C} test split are used as the In-Domain (ID) and Out-Of-Domain (OOD) test dataset, separately. The TriviaQA \cite{triviaqa} train split is used as the training dataset of OEQA task, the TriviaQA dev split and the NQ \cite{nq} dev split are the In-Domain (ID) and Out-Of-Domain (OOD) test dataset. The details of the datasets are listed in Table 8.

\begingroup
\fontsize{9}{11}\selectfont
\setlength{\tabcolsep}{1mm}
\begin{table}[h]
\centering
\begin{tabular}{|c|cc|cc|}
\hline
\textbf{}      & \multicolumn{2}{c|}{\textbf{MCQA}}        & \multicolumn{2}{c|}{\textbf{OEQA}}  \\ \hline
Train       & \multicolumn{1}{l|}{MMLU test}  & 14,079 & \multicolumn{1}{l|}{TriviaQA train} & 87,622 \\ \hline
ID Eval  & \multicolumn{1}{l|}{MMLU val}   & 1,540  & \multicolumn{1}{l|}{TriviaQA dev}   & 11,313 \\ \hline
OOD eval & \multicolumn{1}{l|}{ARC-c dev} & 1,172  & \multicolumn{1}{l|}{NQ dev}         & 3,610  \\ \hline
\end{tabular}
{\fontsize{9pt}{11pt}\selectfont
\caption{Details of the Datasets.}}
\label{table:dataset_details}
\end{table}
\endgroup

\subsection{Prompts}
\subsubsection{Prompts for knowledge state query.}
Prompts for knowledge state queries on MMLU and TriviaQA datasets are shown in Table 9 and Table 10.They use 5-shot and 3-shot settings respectively.

\begin{table}[]
\centering

\begin{tcolorbox}[title={In-Context Examples}, colback=white, coltitle=black, colbacktitle=white!0]
There is a single choice question about \{\textit{Task}\}. Answer the question by replying A, B, C or D.\\
\textbf{Question}: \{\textit{Question1}\} \\
\textbf{A}. \{\textit{Content\_of\_A1}\}\\
\textbf{B}. \{\textit{Content\_of\_B1}\}\\
\textbf{C}. \{\textit{Content\_of\_C1}\}\\
\textbf{D}. \{\textit{Content\_of\_D1}\}\\
\textbf{Answer}: \{\textit{Answer1}\}\\
\\
There is a single choice question about \{\textit{Task}\}. Answer the question by replying A, B, C or D.\\
\textbf{Question}: \{\textit{Question2}\} \\
\textbf{A}. \{\textit{Content\_of\_A2}\}\\
\textbf{B}. \{\textit{Content\_of\_B2}\}\\
\textbf{C}. \{\textit{Content\_of\_C2}\}\\
\textbf{D}. \{\textit{Content\_of\_D2}\}\\
\textbf{Answer}: \{\textit{Answer2}\}\\
\\
$\dots$
\\
There is a single choice question about \{\textit{Task}\}. Answer the question by replying A, B, C or D.\\
\textbf{Question}: \{\textit{Question5}\} \\
\textbf{A}. \{\textit{Content\_of\_A5}\}\\
\textbf{B}. \{\textit{Content\_of\_B5}\}\\
\textbf{C}. \{\textit{Content\_of\_C5}\}\\
\textbf{D}. \{\textit{Content\_of\_D5}\}\\
\textbf{Answer}: \{\textit{Answer5}\}\\
\end{tcolorbox}

\begin{tcolorbox}[title={Instruction}, colback=white, coltitle=black, colbacktitle=white!0]
There is a single choice question about \{\textit{Task}\}. Answer the question by replying A, B, C or D.\\
\textbf{Question}: \{\textit{Question}\} \\
\textbf{A}. \{\textit{Content\_of\_A}\}\\
\textbf{B}. \{\textit{Content\_of\_B}\}\\
\textbf{C}. \{\textit{Content\_of\_C}\}\\
\textbf{D}. \{\textit{Content\_of\_D}\}\\
\textbf{Answer}:
\end{tcolorbox}

\caption{The Prompt Template for Knowledge State Query on MMLU. The Italic \{\textit{text}\} in Curly Braces Represents Variables That Need To be Replaced.}
\label{table:prompt_kq_MMLU}

\end{table}

\begin{table}[]
    \centering
    
    \begin{tcolorbox}[title={In-Context Examples}, colback=white, coltitle=black, colbacktitle=white!0]
    Answer the following question as simple as possible.\\
    \textbf{Question}: \{\textit{Question1}\}\\
    \textbf{Answer}: \{\textit{Answer1}\}\\
    \\
    Answer the following question as simple as possible.\\
    \textbf{Question}: \{\textit{Question2}\}\\
    \textbf{Answer}: \{\textit{Answer2}\}\\
    \\
    Answer the following question as simple as possible.\\
    \textbf{Question}: \{\textit{Question3}\}\\
    \textbf{Answer}: \{\textit{Answer3}\}\\
    \end{tcolorbox}
    
    \begin{tcolorbox}[title={Instruction}, colback=white, coltitle=black, colbacktitle=white!0]
    Answer the following question as simple as possible.\\
    \textbf{Question}: \{\textit{Question}\}\\
    \textbf{Answer}: 
    \end{tcolorbox}
    
    \caption{The Prompt Template for Knowledge State Query on TriviaQA. The Italic \{\textit{text}\} in Curly Braces Represents Variables That Need To be Replaced.}
    \label{table:prompt_kq_triviaqa}
    
    \end{table}

\subsubsection{Prompts for training.}
For the rehearsal training and \textit{Van-Tuning}, we use the \textit{basic} prompt as shown in Table 9 and Table 10 without in-context example. All other experiments use the \textit{refuse} prompt as shown in Table 11 and Table 12. Loss is only computed on the target answer $\{\text{answer}_{rait}\}$.

\begin{table}[]
\centering

\begin{tcolorbox}[title={Instruction}, colback=white, coltitle=black, colbacktitle=white!0]
There is a single choice question about \{\textit{Task}\}. If you know the answer, please directly respond with the correct answer A, B, C, or D. If you do not know the answer, please respond with ``I don't know.".\\
\textbf{Question}:\{\textit{Question}\} \\
\textbf{A}. \{\textit{Content\_of\_A}\}\\
\textbf{B}. \{\textit{Content\_of\_B}\}\\
\textbf{C}. \{\textit{Content\_of\_C}\}\\
\textbf{D}. \{\textit{Content\_of\_D}\}\\
\textbf{Answer}: \{$Answer_{\text{rait}}$\} 
\end{tcolorbox}

\caption{The \textbf{REFUSE} Prompt Template for \textbf{Training} on MMLU. The The Italic \{\textit{text}\} in Curly Braces Represents Variables That Need To be Replaced.}
\label{table:prompt_training_refuse_mmlu}

\end{table}




\begin{table}[]
\centering

\begin{tcolorbox}[title={Instruction}, colback=white, coltitle=black, colbacktitle=white!0]
Answer the following question, and if you don't know the answer, only reply with ``I don't know":\{\textit{Question}\}\\
\{$Answer_{\text{rait}}$\} 
\end{tcolorbox}

\caption{The \textbf{REFUSE} Prompt Template for \textbf{Training} on TriviaQA. The Italic \{\textit{text}\} in Curly Braces Represents Variables That Need To be Replaced.}
\label{table:prompt_training_refuse_triviaqa}

\end{table}

\subsubsection{Prompts for evaluation.}
The \textit{Init-Basic} method uses the original question format for evaluation, without any prior instructions. For the other methods, the evaluation prompts are shown in Tables 16 and 17.

\begin{table}[]
\centering

\begin{tcolorbox}[title={Instruction}, colback=white, coltitle=black, colbacktitle=white!0]
There is a single choice question about \{\textit{Task}\}. If you know the answer, please directly respond with the correct answer A, B, C, or D. If you do not know the answer, please respond with ``I don't know.".\\
\textbf{Question}:\{\textit{Question}\} \\
\textbf{A}. \{\textit{Content\_of\_A}\}\\
\textbf{B}. \{\textit{Content\_of\_B}\}\\
\textbf{C}. \{\textit{Content\_of\_C}\}\\
\textbf{D}. \{\textit{Content\_of\_D}\}\\
\textbf{Answer}: 
\end{tcolorbox}

\caption{The \textbf{REFUSE} Prompt Template for \textbf{Evaluation} on MMLU. The Italic \{\textit{text}\} in Curly Braces Represents Variables That Need To be Replaced.}
\label{table:prompt_eval_refuse_mmlu}

\end{table}




\begin{table}[]
\centering

\begin{tcolorbox}[title={Instruction}, colback=white, coltitle=black, colbacktitle=white!0]
Answer the following question, and if you don't know the answer, only reply with ``I don't know": \{\textit{Question}\}
\end{tcolorbox}

\caption{The \textbf{REFUSE} Prompt Template for \textbf{Evaluation} on TriviaQA. The Italic \{\textit{text}\} in Curly Braces Represents Variables That Need To be Replaced.}
\label{table:prompt_eval_refuse_triviaqa}

\end{table}

\subsection{Training}

\subsubsection{LoRA or full parameter tuning} In the MCQA task, we used LoRA, setting $r=64$, $\alpha=16$, $dropout=0.1$.
In the OEQA task, we used full parameter training. 
\subsubsection{Learning rate} 
We use the cosine learning rate scheduler with warming up. The maximum learning rate for MCQA task is 2e-4. The maximum learning rate on OEQA for LLaMA2-7B-Chat is 2e-5, for LLaMA3-8B-Instruct is 8e-6. 

\subsection{Evaluation}

During the evaluation, the LLM answers questions in the validation set in generation mode. When parsing the results, the algorithm first checks whether the LLM's response is refusal reply. We have constructed a list of refusal keywords, and the algorithm checks whether the response generated by the LLM contains any refusal keywords. If so, it is considered that the model refuses to answer the question; otherwise, it is considered that the LLM has answered normally. Conversely, for questions that the LLM answers normally, we parse the answers by rules and calculate metrics based on the ground truth. Specifically, for the MCQA tasks (MMLU, ARC-c), we parse the LLM's response by rules to obtain the four options ABCD; for the OEQA tasks (TriviaQA, NQ), the answers in the dataset usually contain one or more candidates. If any candidate is included in the LLM's response, it is considered that the LLM has answered correctly; otherwise, it is incorrect.

\end{document}